# Plan-based Policies for Efficient Multiple Battery Load Management

**Maria Fox**                                                    MARIA.FOX@KCL.AC.UK
**Derek Long**                                                   DEREK.LONG@KCL.AC.UK
**Daniele Magazzeni**                                   DANIELE.MAGAZZENI@KCL.AC.UK
*Department of Informatics*
*King's College London*
*Strand, London WC2R 2LS, UK*

## Abstract

Efficient use of multiple batteries is a practical problem with wide and growing application. The problem can be cast as a planning problem under uncertainty. We describe the approach we have adopted to modelling and solving this problem, seen as a Markov Decision Problem, building effective policies for battery switching in the face of stochastic load profiles.

Our solution exploits and adapts several existing techniques: planning for deterministic mixed discrete-continuous problems and Monte Carlo sampling for policy learning. The paper describes the development of planning techniques to allow solution of the non-linear continuous dynamic models capturing the battery behaviours. This approach depends on carefully handled discretisation of the temporal dimension. The construction of policies is performed using a classification approach and this idea offers opportunities for wider exploitation in other problems. The approach and its generality are described in the paper.

Application of the approach leads to construction of policies that, in simulation, significantly outperform those that are currently in use and the best published solutions to the battery management problem. We achieve solutions that achieve more than 99% efficiency in simulation compared with the theoretical limit and do so with far fewer battery switches than existing policies. Behaviour of physical batteries does not exactly match the simulated models for many reasons, so to confirm that our theoretical results can lead to real measured improvements in performance we also conduct and report experiments using a physical test system. These results demonstrate that we can obtain 5%-15% improvement in lifetimes in the case of a two battery system.

## 1. Introduction

In this paper we describe an application of planning to the important problem of multiple battery management. The paper is an extended and developed version of work originally presented at the International Conference on Automated Planning and Scheduling (Fox, Long, & Magazzeni, 2011) and, in particular, adds physical results to the work described in that paper.

An increasing number of systems depend on batteries for power supply, ranging from small mobile devices to very large high-powered devices such as batteries used for local storage in electrical substations. In many of these systems there are significant user-benefits, or engineering reasons, to base the supply on multiple batteries, with load being switched between batteries by a control system. In order to power such systems for the longest time possible, it is necessary to devise switching strategies that extract the maximum possible lifetime out of the batteries. We show how planning is used as the basis of a highly efficient switching strategy.

Due to the physical and chemical properties of batteries, it is possible to extract a greater proportion of the energy stored in a single battery of capacity $C$ than of that stored in $n$ batteries each of capacity $C/n$, for $n > 1$. Throughout this paper, when we refer to the *efficiency* of a switching strategy in the use of multiple batteries, we are talking about the proportion of the charge we extract from the batteries to service a load, compared with servicing the same load from a single battery with capacity equal to the combined collection of batteries and equivalent physical properties. If this proportion is very high, for example: over 90%, then the switching strategy can be considered highly efficient.

The key to efficient use of multiple batteries lies in the design of effective policies for the management of the switching of load between them. We are concerned with the situation in which the load can be serviced entirely by one of a suite of batteries at a time, so that the charge of that battery drains while the other batteries' charge levels remain static. This problem is distinct from the problem of managing cells within a single battery, where the objective is usually to keep the charge in the cells level. Batteries exhibit the phenomenon of *recovery*, which is a consequence of the chemical properties of a battery: as charge is drawn from a battery, the stored charge is released by a chemical reaction, which takes time to replenish the charge. In general, charge will be drawn from a battery faster than the reaction can replenish it and this can lead to a battery appearing to become dead when, in fact, it still contains stored charge. Therefore, more efficient use of multiple batteries can be achieved by exploiting recovery. By allowing the battery to rest, the reaction can replenish the charge and the battery become functional once again. Thus, efficient use of multiple batteries involves carefully timing the use and rest periods. Determining this timing can be seen as a planning problem.

The paper is organised as follows. We begin by presenting the multiple battery usage problem in detail, and describing the battery model we use.

In Section 4 we describe the approach we have adopted for solving the deterministic version of the problem, where we assume that we know the load profile to service. We provide a PDDL+ encoding of the problem and we describe a planning technique for dealing with the continuity involved in the domain. We complete this section by comparing the performance of plan-based solutions with the best policies currently considered for multiple battery management.

In Section 5 we show how the high quality plans obtained for the deterministic problems can be used to learn an efficient policy for the general case where the load profiles are not known in advance. We describe the classification process we have used and we evaluate the performance of the policy when servicing stochastic load profiles. Related work is then discussed in Section 6.

In Section 7 we present the details of a physical experiment, using 6 Volt lead acid batteries, which we conducted in order to confirm our simulation results. We describe the experimental setup and, in the interests of reproducibility, the parameter estimation process we have followed. We then report our experimental results and discuss their significance.

Section 8 outlines our plans for future work and Section 9 concludes the paper.

## 2. Motivations

Many electrically powered systems rely on large, heavy batteries to supply adequate levels of power and current. If the power requirements of these devices can be supplied by multiple lightweight batteries, coordinated to supply the same load as would typically be supplied by a much larger battery, this could significantly change the way these devices are used and the range of applications to which they might be suited.

Examples of powered systems that could benefit from distribution of the battery power include externally powered electric prosthetics. Prostheses powered by electric motors can be more functional and more attractive than body-powered prosthetics, but they can be heavy and expensive. The power requirements of a capable prosthetic arm, combining an elbow with a dexterous hand, necessitate a large, and hence heavy, battery. The high torque motors required to drive a prosthetic elbow require high voltages and current, while modern dexterous hands require significantly more current than did the traditional single-motor electric hands.

While a primitive prosthetic arm could run both the elbow and the hand on a 1 Amp Hour battery, dexterous hands require batteries with as much as 2 Amp Hour capacities, and if the hand and the elbow are to be run off the same battery, then even more current and larger capacities are needed with a consequent increase in weight and heat. The high power demand requires that either multiple batteries are carried or batteries are frequently recharged or replaced. The weight of externally powered prostheses is a common source of dissatisfaction amongst users and the placement of batteries to minimise the weight effects is an important part of the prosthetic design. If the battery power can be distributed around the body, with the power requirements being met by carefully coordinated multiple independent batteries of the same power but much smaller capacity, then the weight issue can be made less significant to the user, and the heat generated by the batteries can also be reduced making them more comfortable to wear.

The same benefits can potentially be obtained in any situation where batteries have to be carried in order to power portable electrical devices. Military personnel currently carry about 20kg of batteries into the field to power their communication equipment, vision and sensing systems and other electronic devices. Robotic devices are often battery powered and rely on carrying large numbers of batteries to maximise operational lifetime. Electric cars typically carry multiple batteries, although they must sometimes be used in series to maximise power availability. This creates different constraints on the way they can be used from those we consider in this paper. However, as the technology develops, opportunities will arise for exploiting partitioned batteries in electric vehicles.

One of the advantages of being able to distribute battery power across multiple independent batteries is the ability to swap batteries out as they die, requiring a few small battery spares to be carried instead of one large one. This "hot-swapping" capability could have an important role to play in mobile computing devices where, instead of having to recharge the battery every 6 hours or so, continuous power over a longer period could be achieved by selectively replacing spent cells.

The major motivation for the work we have done is therefore to obtain close-to-optimal battery performance for high-powered devices, while benefitting from the ability to distribute the weight and heat production.

## 3. The Multiple Battery Usage Problem

The multiple battery usage planning problem has been explored by several authors, from an electrical engineering perspective, for example in the work of Benini et al. (2003) and Rao et al. (2003), and also from a scheduling perspective (Jongerden, Haverkort, Bohnenkamp, & Katoen, 2009) and an optimisation perspective (Wang & Cassandras, 2011) (in the latter, the simplifying assumption that load can be shared arbitrarily between batteries is made). Benini et al. construct a very accurate battery model, parameterising it to capture lithium-ion, cadmium-nickel and lead-acid battery types, and show how hand constructed policies can achieve efficiency, relative to a single battery, between 70% and 97.5%. To achieve this, the policy is constructed to select a new battery whenever the voltage of the battery currently servicing a load drops below a certain threshold. The next battery is selected according to one of four alternative policies (Benini et al., 2003):

- $V_{max}$: select the battery pack with highest state of charge.

- $V_{min}$: select the battery pack with lowest state of charge.

- $T_{max}$: select the battery pack that has been unused for the longest time.

- $T_{min}$: select the battery that has been unused for the shortest time.

The authors show that $V_{max}$ is the best of these policies, tested on up to four batteries. In the general case of $n$ batteries, the $V_{max}$ is referred to as *best-of-n*.

Jongerden et al. (2009) uses a model checking strategy, based on UPPAAL, to schedule battery use given a known load profile. The approach is based on the use of a different battery model, the Kinetic Battery Model, discussed in more detail below. This is a non-linear continuous model and the authors treat it by discretisation and scheduling to a horizon. This approach allows them to find highly effective schedules, but it does not scale well because of the need to use a fine-grained discretisation of the temporal dimension. It is worth emphasising, since it contrasts with our approach, that Jongerden et al. work with a fixed size discretisation of time, allowing them to focus on scheduling the resources (batteries) into the load periods.

In deployed systems, the standard policies are typically static, based on rapid switching between available batteries. In fact, an optimal use of multiple batteries can be achieved theoretically by switching between them at extremely high frequency, when the behaviour converges on that of a single battery (Rao et al., 2003). Unfortunately, this theoretical solution is not achievable in practice because of the losses in the physical process of switching between batteries, as the frequency increases. In fact, switching losses in MOSFETs are approximately linearly dependent on switching frequency and also on the current being switched (Eberle, 2008). $T_{max}$ and $V_{max}$ policies applied at fixed frequencies are the most commonly fielded solutions, but these often achieve less than 80% efficiency (Benini et al., 2003).

## 3.1 Objectives

In this paper our objective is to construct policies for multiple battery problems, where load is modelled probabilistically using known distributions for load size, load duration and load frequency (or equivalently, the gaps between successive loads). Our primary purpose, in constructing these policies, is to achieve the longest possible battery lifetime. The best deployed solutions typically deliver less than 80% efficiency, while the best published solutions deliver less than about 95% efficiency (our reading suggests that these high values are in simulation rather than in physical experiments). We show that our approach, based on construction of optimising solutions to Monte Carlo sampled problem instances and their use in the construction of appropriate policies, produces robust solutions that deliver better than 99% efficiency in simulation. Furthermore, as a side-effect of the way in which these solutions are constructed, we achieve this efficiency in lifetime while using smaller numbers of battery switches than published policies. This beneficial side-effect reduces the potential switching losses in implementing the policy. We use the Kinetic Battery Model (Manwell & McGowan, 1993) (KiBaM) as the basis of our construction of optimising solutions and this raises challenges in the treatment of the non-linear mixed discrete-continuous optimisation problem, as we discuss below.

### 3.2 The Kinetic Battery Model

In the Kinetic Battery Model (Manwell & McGowan, 1993; Jongerden et al., 2009) the battery charge is distributed over two wells: the available-charge well and the bound-charge well (see Figure 1).

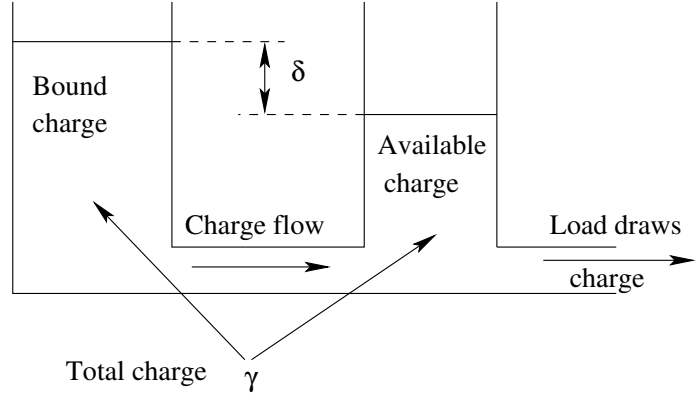

Figure 1: Kinetic Battery Model

A fraction $c$ of the total charge is stored in the available-charge well, and a fraction $1 - c$ in the bound-charge well. The available-charge well supplies electrons directly to the load ($i(t)$), where $t$ denotes the time, whereas the bound-charge well supplies electrons only to the available-charge well. The charge flows from the bound-charge well to the available-charge well through a "valve" with fixed conductance, $k$. Moreover, the rate at which charge flows between the wells depends on the height difference between the two wells. The heights of the two wells are given by:

$$h_1 = \frac{y_1}{c} \qquad h_2 = \frac{y_2}{1-c}$$

where $y_1$ is the the available charge and $y_2$ is the bound charge. When a load is applied to the battery, the available charge reduces, and the height difference between the two wells grows. When the load is removed, charge flows from the bound-charge well to the available-charge well until the heights are equal again. The change in the charge in both wells is given by the following system of differential equations:

$$\begin{cases} \frac{dy_1}{dt} = -i(t) + k(h_2 - h_1) \\ \frac{dy_2}{dt} = -k(h_2 - h_1) \end{cases}$$

with initial conditions $y_1(0) = c \cdot C$ and $y_2(0) = (1-c) \cdot C$, where $C$ is the total battery capacity.

To describe the discharge process of the battery, as in Jongerden et al. (2009), we adopt coordinates representing the height difference between the two wells, $\delta = h_2 - h_1$, and the total charge in the battery, $\gamma = y_1 + y_2$. In this new setting $y_1 = c(\gamma - (1-c)\delta)$.

The change in both wells is then given by the system of differential equations

$$\begin{cases} \frac{d\delta}{dt} = \frac{i(t)}{c} - k'\delta \\ \frac{d\gamma}{dt} = -i(t) \end{cases}$$

with solutions

$$\begin{cases} \delta(t) = \frac{i}{c} \cdot \frac{1 - e^{-k't}}{k'} \\ \gamma(t) = C - it \end{cases}$$

where $k' = k/(1-c)c$, $\delta(0) = 0$ and $\gamma(0) = C$. The condition for a battery to be empty is $\gamma(t) = (1-c)\delta(t)$.

This model is less sophisticated than that used by Benini et al. (2001), but a comparison of battery models by Jongerden and Haverkort (2009) concludes that the Kinetic Battery Model (KiBaM) is the best for performance modelling.

### 3.3 Battery Usage Planning

Although the battery load management can be seen as a scheduling problem, the setting we consider makes it a planning problem. For a given a load profile to service, if we knew the number of switching actions between batteries that would be required, but not the times at which these actions should be performed, then the problem could be managed as a scheduling problem. In our case, however, the number of switching actions cannot be identified in advance, as each period of load can be shared arbitrarily between different batteries. Thus, the battery load management becomes a planning problem. By discretising time to the shortest time over which a battery must be in use, it is possible to construct a scheduling problem in which the maximum possible number of battery switches is considered, where some of the switches might not be used. The difficulty in this approach is that the shortest period of use can be very short compared with the battery lifetime: in our physical experiments (Section 7), for example, the maximum number of switches would be over 700, while for larger capacity batteries or smaller loads the number of switches could easily be several thousand. The scheduling approach used by Jongergen et al. (2009) cannot scale to manage more than a few tens of intervals.

Furthermore, the KiBaM, which is a deterministic non-linear continuous model of battery performance, lends itself, in principle, to use in an optimisation problem solver that can find the best battery usage plan, given a load profile. The multiple battery usage problem, in its deterministic form, is clearly an optimisation problem and Wang and Cassandras (2011) have shown that, under certain assumptions, it can be tackled analytically (despite being non-linear), using the KiBaM. In order to do so they assume that load can be split arbitrarily between batteries (which is not easily achievable in practice). They also assume that the load can be serviced in an arbitrary schedule within a given timespan, provided that the total charge drawn from the batteries meets a required workload. This second assumption is not consistent with our own situation, in which load must be serviced according to demands placed by a user at specific times, without flexibility. Unfortunately, their analysis cannot be modified to deal with the situation we consider.

It is of interest to speculate on whether a standard Operations Research approach, using some form of Mixed-Integer Linear Program (MILP) model, might be used to solve the deterministic multiple battery usage problem. At first glance the answer is trivial: since the model is non-linear, it is clear that a MILP cannot be used. A more sophisticated approach might be considered, using an approximation of the exponential recovery curves using piece-wise linear components. However, because the precise shape of these recovery curves depends on the state of charge of the battery at the start of the period of recovery (both its available and bound parts), the approximations must either be built dynamically, or else the model must anticipate all possible states of charge at all times points, effectively building the entire search space of the states of charge of the battery into the model. The former approach cannot be achieved in a standard MILP and we are not aware of any solving technology that could manage this approach; the latter approach is obviously impractical for anything but the most trivial of situations.

In most real battery usage problems the load profile is generated by external processes, typically controlled directly or indirectly by user demands. These demands can often be modelled probabilistically, reflecting typical patterns of use. In our work we assume that the profiles are drawn from a known distribution. The consequence is that the planning problem ceases to be a deterministic optimisation problem, but a probabilistic problem in which the plan must be a policy, as discussed in Section 5.

### 3.4 Our Approach

We adopt an approach based on a combination of two ideas. Firstly, we sample from the distribution of loads to arrive at a deterministic problem, which we then solve using the continuous KiBaM as our battery model. This leads to an interesting continuous non-linear optimisation problem, which we solve using a discretise-and-validate approach. Currently we are using UPMurphi (Della Penna, Intrigila, Magazzeni, & Mercorio, 2009) to solve the deterministic instances but, after discretisation, any metric temporal planner could be used in principle. Secondly, we use a decision tree classifier to combine the solutions to the sample problem instances and learn a policy for the MDP from which the problems are drawn. The classification process maps states into actions and produces a policy in the form of a decision tree.

Our approach is domain-specific in some respects:

- Our discretisation scheme, while based on general principles, is selected for the problem domain and load distribution.

- We use a search heuristic that, while not restricted to the battery problem alone, is not suited to all problems.

- The aggregation of solutions into a policy makes use of an entirely general approach, but the extent to which the approach yields good policies will depend on the nature of the problem space in which it is applied.

We make use of existing tools as far as is possible, to simplify the construction of our solution.

## 4. Solving Deterministic Multiple Battery Problems

In this section we consider the multiple battery management problem as an optimisation problem, when faced with a known and deterministic load profile.

### 4.1 A PDDL+ Battery Model

PDDL+ (Fox & Long, 2006) is an extension of the standard planning domain modelling language, PDDL, to capture continuous processes and events. The dynamics of KiBaM can be captured very easily in PDDL+. In Figure 2 we show the two processes, `consume` and `recover`, that govern the behaviour of batteries and the event triggered by attempting to load a battery once its available charge is exhausted. In addition, there is a durative action of variable duration that allows the planner to use a battery over an interval (see Figure 3). The two processes are active whenever their preconditions are satisfied, meaning that they usually execute concurrently. Together, they model both the draining of charge and the recovery that are described in the differential equation $d\delta/dt$. An event is triggered if there is ever a positive load and no active service.

```
(:process consume
 :parameters (?b - battery)
 :precondition (switchedOn ?b)
 :effect (and (decrease (gamma ?b) (* #t (load)))
              (increase (delta ?b) (* #t (/ (load) (cParam ?b)))))
)

(:process recover
 :parameters (?b - battery)
 :precondition (>= (delta ?b) 0)
 :effect (and (decrease (delta ?b) (* #t (* (kprime ?b) (delta ?b)))))
)

(:event batteryDead
 :parameters (?b - battery)
 :precondition (and (switchedOn ?b)
                    (<= (gamma ?b) (* (-1 (cParam ?b)) (delta ?b))))
 :effect (and (not (switchedOn ?b)) (dead ?b))
)
```

Figure 2: Part of PDDL+ encoding of KiBaM dynamics

```
(:durative-action use
:parameters (?b - battery)
:duration (>= ?duration 0)
:condition (and (at start (switchedOff ?b))
               (over all (switchedOn ?b)))
:effect (and (at start (and (switchedOn ?b) (not (switchedOff ?b))
                            (increase (services) 1)))
            (at end (and (switchedOff ?b) (not (switchedOn ?b))
                         (decrease (services) 1))))
)
```

Figure 3: PDDL+ durative action for battery use

The load profile to be serviced is encoded in the PDDL+ problem through the use of timed initial literals, which allow expression of exogenous events corresponding, in our case, to changes in the load value. A fragment of the problem (which also contains the battery specification) is shown in Figure 4.

The use of PDDL+ as our modelling language grants several benefits. Firstly, it allows us to use VAL (Howey, Long, & Fox, 2004) to validate solutions analytically against the continuous model, allowing us to confirm that the discretisation we use during construction of solutions does not compromise the correctness of the plan. Secondly, it provides us with a semantics for our model in terms

```
(define (problem 2B) (:domain kibam)
(:objects b1 b2 - battery)
(:init
(= (cParam b1) 0.166)
(= (kprime b1) 0.122)
(= (gamma b1) 5.5)
(= (delta b1) 0)
...
(at 0 (= (load) 0.25))
(at 1.00 (= (load) 0.50))
(at 2.00 (= (load) 0.25))
(at 3.00 (= (load) 0.50))
(at 4.00 (= (load) 0.25))
...
```

Figure 4: Fragment of the PDDL+ problem

of a timed hybrid automaton as described by Fox and Long (2006). Finally, we can make use of existing tools that construct and search in spaces defined by PDDL+ models, such as UPMurphi (Della Penna et al., 2009).

In their paper on PDDL+, Fox and Long (2006) propose a semantics based on a mapping to timed hybrid automata (Alur & Dill, 1994). The semantics of the domain instantiated for two batteries is given by the three hybrid automata shown in Figure 5, where variables d, g, L and s refer to PDDL+ functions delta, gamma, load and services, respectively. This semantics is one route by which model-checking systems designed to manage timed hybrid automata can be adapted to operate directly on the battery problem. The batteries reveal their non-linear behaviour in the definitions of the expressions governing the rates of change of both $d_1$ and $d_2$ in the pair of states *switchedOnB1* and *switchedOffB1* and the equivalent pair for $B2$. Unfortunately, these equations are beyond the reach of most current model-checking systems, but by discretising the ranges of these variables the functions can be managed by UPMurphi.

The variable T is the *time-slip* variable introduced by Fox and Long (2006) which allows the correct modelling of PDDL+ domains with events in standard hybrid automata. In particular, the time-slip variable increases at rate 1 whenever the preconditions of the events disaster (positive load and no battery being used) or notOptimal (a battery being used without any load to service) are satisfied. Each state in the three hybrid automata has an invariant condition stating that the time-slip variable must be 0, and this guarantees that the events will be applied as soon as their preconditions become true, without any action transitions occurring between.

## 4.2 The Discretise-and-Validate Approach

Our technique is based on a discretise-and-validate approach (see Figure 6), in which the continuous dynamics of the problem are relaxed into a discretised model, where discrete time steps and corresponding step functions for resource values are used in place of the original continuous dynamics. This relaxed problem is solved using a forward reachability analysis and then solutions are validated against the continuous model using the validator, VAL (Howey et al., 2004), which provides analytic solutions to differential equations involved in the models.

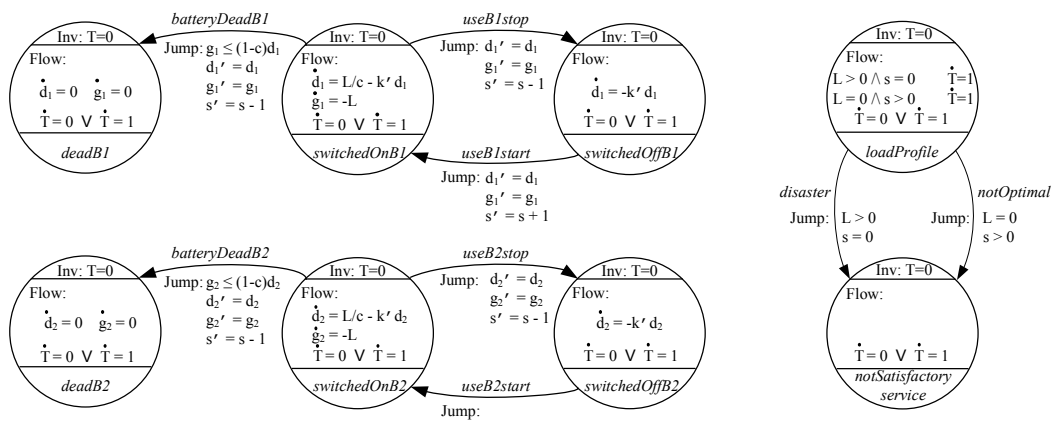

Figure 5: Hybrid automata modelling two kinetic batteries scheduling

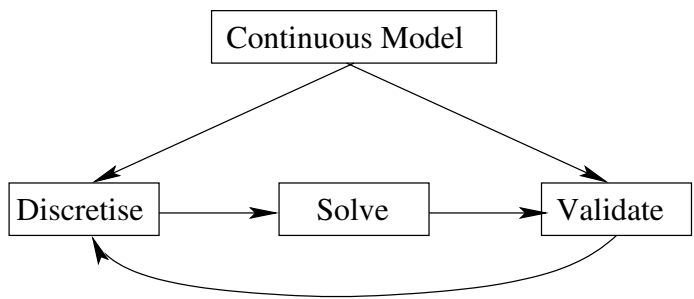

Figure 6: The Discretise and Validate Approach

The validation process is used to identify whether a finer discretisation is required and guide remodelling of the relaxed problem. As an example, in our simulation, we first considered a time discretisation $\delta_t = 0.1$, and obtained the plan shown in Figure 7 (left). However, when we validated the discrete solution generated by the planner against the continuous model, we found out that the solution was indeed not valid, as highlighted in the following fragment of the VAL report:

```
Checking next happening (time 5.08986)
Updating (gamma b1) (0.502404) by 0.337447 assignment
Updating (delta b1) (0.328362) by 0.550475 assignment
Updating (delta b2) (0.405504) by 0.257052 assignment

EVENT triggered at (time 5.08986)
Triggered event (batterydead b1)
Deleting (switchedon b1)
Adding (dead b1)

Invariant for (use b1) has its condition unsatisfied
between time 5.08986 to 5.1.
```

```
0.0:   (use b1)      [3.40]          0.0:   (use b1)      [3.40]
3.40:  (use b2)      [0.50]          3.40:  (use b2)      [0.50]
3.90:  (use b1)      [0.10]          3.90:  (use b1)      [0.10]
4.00:  (use b2)      [0.50]          4.00:  (use b2)      [0.50]
4.50:  (use b1)      [0.60]          4.50:  (use b1)      [0.58]
5.10:  (use b2)      [0.20]          5.08:  (use b2)      [0.27]
5.30:  (use b1)      [0.30]          5.35:  (use b1)      [0.08]
5.60:  (use b2)      [0.20]          5.43:  (use b2)      [0.57]
5.80:  (use b1)      [0.30]          6.10:  (use b1)      [0.05]
6.10:  (use b2)      [0.70]          6.15:  (use b2)      [0.40]
6.80:  (use b1)      [0.20]          6.55:  (use b1)      [0.05]
7.00:  (use b2)      [1.30]          6.60:  (use b2)      [0.50]
8.30:  (use b1)      [0.20]          7.10:  (use b1)      [0.05]
8.50:  (use b2)      [0.20]          7.15:  (use b2)      [1.00]
8.70:  (satisfied)                   8.15:  (use b1)      [0.30]
                                     8.45:  (use b2)      [0.20]
                                     8.65:  (use b1)      [0.05]
                                     8.70:  (satisfied)
```

Figure 7: Plans generated using different time discretisations: $\delta_t = 0.1$ (left) and $\delta_t = 0.01$ (right)

The very precise analysis provided by VAL allows us to know the exact value of the charge in the (simulated) batteries during the execution of the plan. In this example, the charge in battery 1 terminates 0.01014 time units before the time expected with the discretised model. This suggests a refinement of the discretisation, setting $\delta_t = 0.01$, which eventually produced a valid plan, shown in Figure 7 (right). As can be seen, the finer discretisation handles very sensitive interactions and the system switches to battery 2 when charge in battery 1 is almost fully drained (at time point 5.08).

Although Jongerden et al. (2009) also use a discretisation approach, they fix the granularity of the time-step in advance. In contrast, we use a variable sized discretisation, by allowing a range of alternative step sizes to be considered during search.

We now introduce the formal statement of the deterministic version of the problem we are interested in. A *hybrid system* is a system whose state description involves continuous as well as discrete variables. We approximate the system by discretising the continuous components of the state (which we assume to be bounded) and their dynamic behaviours so obtaining a finite number of states.

**Definition 1 (Finite State Temporal System)** *A* Finite State Temporal System *(FSTS)* $\mathcal{S}$ *is a 5-tuple* $(S, s_0, \mathcal{A}, \mathcal{D}, F)$, *where:* $S$ *is a finite set of* states, $s_0 \in S$ *is the* initial state, $\mathcal{A}$ *is a finite set of* actions, $\mathcal{D}$ *is a finite set of* durations *and* $F : S \times \mathcal{A} \times \mathcal{D} \to S$ *is the* transition function, *i.e.* $F(s, a, d) = s'$ *iff the system can reach state $s'$ from state $s$ via action $a$ having a duration $d$. For each state $s \in S$, we also define the set* EnAct(s)= $\{a \in \mathcal{A} | \exists d \in \mathcal{D} : F(s, a, d) \in S\}$, *as the set of all the actions* enabled *at state $s$.*

In an FSTS, each state $s \in S$ is assumed to contain a special *temporal variable* $t$ denoting the time elapsed in the current path from the initial state to $s$. In the following we use the notation $t(s)$ for the value of variable $t$ in state $s$. For all $s_i, s_j \in S$ such that $F(s_i, a, d) = s_j$, $t(s_j) = t(s_i) + d$.

**Definition 2 (Trajectory)** A trajectory *in the FSTS* $\mathcal{S} = (S, s_0, \mathcal{A}, \mathcal{D}, F)$ *is a sequence* $\pi = s_0 a_0 d_0 s_1 a_1 d_1 s_2 a_2 d_2 \ldots s_n$ *where,* $\forall i \geq 0$, $s_i \in S$ *is a state,* $a_i \in \mathcal{A}$ *is an action,* $d_i \in \mathcal{D}$ *is a duration and* $F(s_i, a_i, d_i) = s_{i+1}$. *If* $\pi$ *is a trajectory, we write* $\pi_s(i)$, $\pi_a(i)$ *and* $\pi_d(i)$ *to denote the state* $s_i$, *the action* $a_i$ *and the duration* $d_i$, *respectively. Finally, we denote with* $|\pi|$ *the length of* $\pi$, *given by the number of actions in the trajectory, and with* $\tilde{\pi}$ *the duration of* $\pi$, *i.e.* $\tilde{\pi} = \sum_{i=0}^{|\pi|-1} \pi_d(i)$.

In order to define the planning problem for such a system, we assume that a set of *goal states* $G \subseteq S$ has been specified. Moreover, to have a finite state system, we fix a *finite temporal horizon, T,* and we require a plan to reach the goal within time $T$. In the case of the battery usage planning problem, this horizon is very important because it represents the target duration for the service provided by the battery. In fact, a good upper bound can be found for the battery problem, which is discussed further in section 4.3.

**Definition 3 (Planning Problem on FSTS)** *Let* $\mathcal{S} = (S, s_0, \mathcal{A}, \mathcal{D}, F)$ *be an FSTS. Then, a* planning problem *(PP) is a triple* $\mathcal{P} = (\mathcal{S}, G, T)$ *where* $G \subseteq S$ *is the set of the goal states and* $T$ *is the finite temporal horizon. A solution for* $\mathcal{P}$ *is a trajectory* $\pi^*$ *in* $\mathcal{S}$ *s.t.:* $|\pi^*| = n$, $\tilde{\pi}^* \leq T$, $\pi_s^*(0) = s_0$ *and* $\pi_s^*(n) \in G$.

The constraints we add to the temporal planning problem are parameterised and can be iteratively relaxed in order to explore successively larger spaces for plans. We use a finite collection of possible durations for segments of processes (Definition 2). This set can be refined by the addition of smaller durations if successive searches fail to find a solution. Allowing different durations within the same search enables the planner to construct states that interact with executing processes at different time points, while stepping quickly along the timeline where there are no interesting features.

### 4.3 The Monotonicity Property and Planning

The battery domain has an important property that supports a simple heuristic evaluation function for states: the charge in the battery monotonically decreases over time and the optimal solution is the one that gives the longest possible plan. An upper bound on the duration of the solution can be found using the observation that the optimal duration cannot exceed that of a single battery with combined capacity equal to the sum of the capacities of the multiple batteries (assuming the same discharging and flow behaviours). Once we have a horizon, we construct and search our discretised search space. To make this approach practical, it is essential that we have an informed heuristic to search the space. For this domain, duration of the plan to the current state plus total remaining charge is admissible, but completely uninformative, while duration plus total *available* charge is highly informative. This is also equivalent to minimising the total bound charge.

This heuristic is suitable for a class of domains: in any domain where there is a monotonically decreasing resource, and the longest plan is required (such as the satellite domain against a finite amount of resources), a heuristic that sums plan duration and available resource will be informative.

We then use a variant of the best-first search (Algorithm 1) to efficiently explore the reachable space. To use variable discretisation efficiently, we break the symmetry in the structure of the search space that arises from the possible orderings of different length action instances. Redundancy is eliminated by disallowing the use of long duration actions immediately following shorter duration versions of the same actions. Long duration actions can only be used if an event or other action has intervened since the last short action in the family. We also disallow the repeated consecutive use of short duration actions beyond the accumulated duration of the next longer duration action. The longest duration action can be repeated arbitrarily often.

---

**Algorithm 1** Dynamic State Space Search ($\mathcal{P}$)

**Input:** a planning problem $\mathcal{P} = ((S, s_0, \mathcal{A}, \mathcal{D}, F), G, T)$

**Output:** a valid plan $\pi^*$

 1: $Q \leftarrow (s_0, \texttt{null}, 0)$;
 2: $H \leftarrow s_0$;
 3: **if** $s_0 \in G$ **then return** $\pi^*$;
 4: **while** $Q \neq \emptyset$ **do**
 5: $\quad (s_h, a_i, d_k) \leftarrow \text{argmax}_{(s,a,d) \in Q} h(s)$;
 6: $\quad$ **for all** $a_j \in \texttt{EnAct}(s_h)$ **do**
 7: $\quad\quad$ **if** $a_j \neq a_i$ **then** $\Delta \leftarrow \{d_l \in \mathcal{D} | t(s_h) + d_l \leq T\}$;
 8: $\quad\quad$ **else** $\Delta \leftarrow \{d_l \in \mathcal{D} | d_l \leq d_k \wedge t(s_h) + d_l \leq T\}$;
 9: $\quad\quad$ **for all** $d_l \in \Delta$ **do**
10: $\quad\quad\quad s' \leftarrow F(s_h, a_j, d_l)$;
11: $\quad\quad\quad$ **if** $s' \in G$ **then return** $\pi^*$;
12: $\quad\quad\quad$ **if** $s' \notin H$ **then**
13: $\quad\quad\quad\quad Q \leftarrow Q \cup (s', a_j, d_l)$;
14: $\quad\quad\quad\quad H \leftarrow H \cup s'$;

---

### 4.4 Plan Search with Variable Discretisation

We now illustrate the way in which the range of differently sized duration intervals can lead to significant benefits in the size of the set of visited nodes in the search space, compared with using a fixed duration increment.

Consider the load profile shown at the top of Figure 8. The planning problem for two batteries is defined according to definitions 1 and 3, with $G = \{s \in S | t(s) = 2.42\}$, i.e. the goal is to service the whole load profile. The temporal horizon $T$ is set to the duration of the profile as well. The definition of the FSTS is straightforward: the set of actions is $\mathcal{A} = \{\texttt{useB1}, \texttt{useB2}, \texttt{wait}\}$ where the former actions refer to the battery being used while the latter one is applicable when there is no active service. The set of durations we use for this example is $\mathcal{D} = \{0.01, 0.4, 0.5, 1.0\}$ (measured in minutes). In practice, to define the set of durations we start with a minimum value and then we add exponentially increasing values up to a maximum duration given by the longest interval between different events (i.e., load variations). In particular, the smallest duration is included in order to handle very sensitive interactions.

In the initial state $s_0$ there is no load and no active service and both batteries have a limited initial capacity. In this setting, the plan search with variable discretisation proceeds as follows:

1. No battery is used for a period of 1 minute (when the load is idle). The corresponding transition is shown in Figure 8.

2. After one minute a load is applied and battery 1 is used. This corresponds to transition $< s_1, \texttt{useB1}, 1.0, s_2 >$. However, for sake of simplicity, let us assume that, due to their limited capacity, batteries cannot be used continuously for 1 minute. The transition is thus not valid and a shorter duration has to be considered.

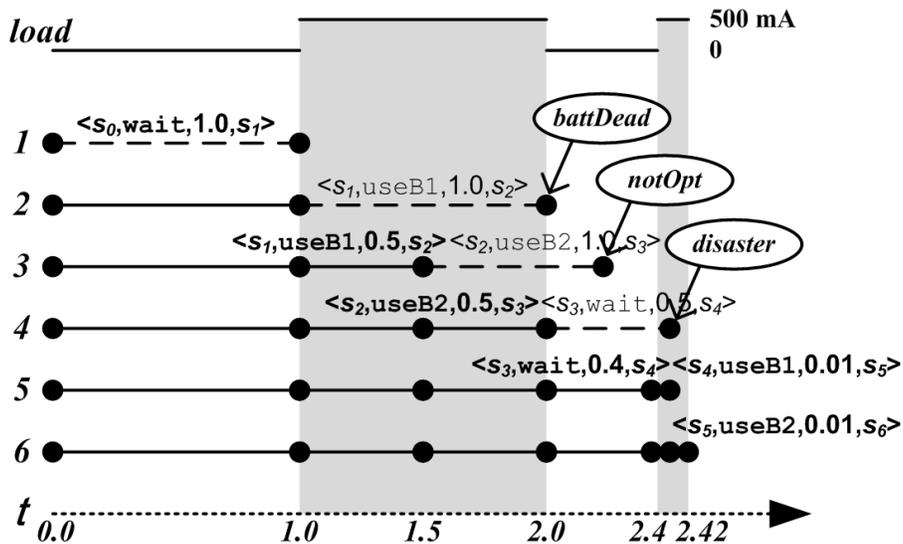

Figure 8: Example of search using variable discretisation

3. Battery 1 is used for 0.5 minute. Then, since a load is still applied, the second battery is used. As before, the transition $< s_2, \texttt{useB2}, 1.0, s_3 >$ can be considered, but in this case there would be an active service and no load.

4. Battery 2 is used for 0.5 minute. In the next period no load is applied, then no battery is used. The transition $< s_3, \texttt{wait}, 0.5, s_4 >$ is considered, but it would lead to a positive load and no active service, so the duration of action $\texttt{wait}$ has to be reduced to 0.4.

5. To service the last load period of 0.02 minute, battery 1 could be used. However, in this sample instance let us assume that the remaining charge in battery 1 allows it to service only 0.01 minute. So, finally, battery 2 is used until the end of the load profile.

The validity of a transition is dynamically checked during the search since invalid transitions trigger specific events (e.g. event $\texttt{batteryDead}$ is triggered at step 2 and event $\texttt{disaster}$ is triggered at step 4) which, in turn, violates the invariant conditions of corresponding actions (a battery must not die during use). Moreover, with variable discretisation only 6 states have to be visited in order to reach the goal, while using a uniform discretisation it is necessary to explore at least 242 states since the finest discretisation of 0.01 must be used in order to correctly handle the interactions in steps 5 and 6.

A further benefit of the use of differently sized durations in the discretisation is that favouring longer durations reduces the number of switches in the solutions we generate, leading to solutions that are better in practical terms than those based on a high frequency switching between batteries, as is shown in subsequent results.

### 4.5 Performance on Deterministic Load Problems

We now present a first set of experimental results to show, in simulation, the performance of our solver on the deterministic battery usage optimisation problem. We use the same case study proposed by Jongerden et al. (2009), where two types of jobs are considered, a low current job (250 mA) and a high current job (500 mA), according to the following load profiles:

- *continuous* loads: one load with only low current jobs (CL_250), one with only high current jobs (CL_500) and one alternating between a low current job and a high current job (CL_alt);

- *intermittent* loads with *short idle periods* of one minute between the jobs: one with only low current jobs (IL$s$_250), one with only high current jobs (IL$s$_500), and one alternating between a low current job and a high current job (IL$s$_alt);

- *intermittent* loads with *long idle periods* of two minutes between the jobs: one with only low current jobs (IL$l$_250) and one with only high current jobs (IL$l$_500).

As a first step, we used these load profiles to validate our variable-range discretisation KiBaM model (planning-KiBaM), and to find an appropriate discretisation for the continuous variables involved in the system dynamics (i.e. variables $\gamma$ and $\delta$ and process durations). To do this we used VAL to validate solutions for the discretised model against the continuous model. As in the work by Jongerden et al. (2009), we considered two battery types, one with capacity 5.5 Amin ($B_1$) and one with capacity 11 Amin ($B_2$). These are small batteries, typical of the capacities of those in small portable devices such as PDAs or mobile phones. Both battery types have the same parameters: $c = 0.166$ and $k' = 0.122\text{min}^{-1}$. We discretised $\gamma$ and $\delta$, rounding them to 0.00001, and, for all the load profiles above and for both battery types, we obtained the same lifetimes computed with the original KiBaM and validated by Jongerden and Haverkort (2008).

To generate the scheduling plans for multiple batteries, we used the approach described in sections 4.2 and 4.3 and the set of durations $\mathcal{D} = \{0.01, 0.02, 0.05, 0.1, 0.25, 0.5, 1.0\}$.

An example of PDDL+ plan is shown in Figure 9, where each row $< t_i, a_i, d_i >$ contains the time point $t_i$ in which action $a_i$ (whose duration is $d_i$) is applied.

```
  0.0:  (use b1)  [1.00]
  1.20: (use b1)  [0.10]
  1.30: (use b2)  [0.10]
  1.80: (use b1)  [0.20]
  2.40: (use b1)  [0.10]
  2.50: (use b2)  [0.10]
  3.10: (use b1)  [1.00]
  4.60: (use b1)  [0.10]
  4.70: (use b2)  [0.10]
  6.20: (use b1)  [0.30]
```

Figure 9: Fragment of the PDDL+ plan

Figure 10 shows a fragment of the corresponding VAL report. Note that VAL provides analytic solutions to the differential equations involved in the KiBaM dynamics.

To evaluate the efficiency of our approach, we compared our solutions to those obtained using the UPPAAL-based approach. The resulting lifetimes are shown in Table 1 where the 'upper bound'

**4.7:** **Event triggered!**
  *Unactivated process* (consume b1)
  *Activated process* (consume b2)

**4.8:** Checking Happening... ...OK!

**4.8:** Checking Happening... ...OK!
  **(delta b1)**$(t) = 2.74431e^{-0.122t}$
  **(gamma b2)**$(t) = -0.3t + 5.44$
  **(delta b2)**$(t) = -14.5542e^{-0.122t} + 14.8134$
  Updating **(delta b1)** (2.74431) by 2.71104 for continuous update.
  Updating **(gamma b2)** (5.44) by 5.41 for continuous update.
  Updating **(delta b2)** (0.259121) by 0.435604 for continuous update.

**4.8:** Checking Happening... ...OK!
  Deleting (switchedon b2)
  Adding (switchedoff b2)
  Decreasing **(services)** (1) by 1.
  Updating **(load)** (0.3) by 0 assignment.

Figure 10: Fragment of VAL report

column shows the theoretical upper bound given by a best-of-two policy with an extremely high-frequency switching. It can be seen, in the first two rows of this table, that the power that can be extracted from a battery with a nominal capacity of 5.5 Amin is only 12.16 min × 250 mA, which is 3.04 Amin, when loading continuously at 250 mA, or 4.59 × 500 mA which is 2.3 Amin when drawing a continuous load of 500 mA. This gives an indication of the extent to which the limit on the conversion of bound charge to available charge affects the performance of batteries.

| load profile | Upper bound lifetime | | Uppaal-KiBaM lifetime | | Planning-KiBaM lifetime (visited states) | |
|---|---|---|---|---|---|---|
| | $B_1$ | $B_2$ | $B_1$ | $B_2$ | $B_1$ | $B_2$ |
| CL_250 | 12.16 | 46.92 | 12.04 | N/A | **12.14** (194) | **46.91** (691) |
| CL_500 | 4.59 | 12.16 | 4.58 | N/A | **4.59** (116) | **12.14** (194) |
| CL_alt | 7.03 | 21.26 | 6.48 | N/A | **7.03** (136) | **21.2** (350) |
| ILs_250 | 44.79 | 132.8 | 40.80 | N/A | **44.76** (552) | **132.7** (1068) |
| ILs_500 | 10.82 | 44.79 | 10.48 | N/A | **10.8** (131) | **44.76** (552) |
| ILs_alt | 16.95 | 72.75 | 16.91 | N/A | **16.92** (159) | **72.55** (599) |
| IL*l*_250 | 84.91 | 216.9 | 78.96 | N/A | **84.88** (488) | **216.8** (1123) |
| IL*l*_500 | 21.86 | 84.91 | 18.68 | N/A | **21.85** (173) | **84.88** (488) |

Table 1: System lifetime (in minutes) for all load profiles according to different battery usages

In all load profiles considered we observe that our approach outperforms the Uppaal-based one significantly, providing solutions that achieve more than 99% efficiency compared with the theoretical limit. The key points described in the preceding parts of this section allow the resulting search to efficiently prune the state space and quickly find the solutions. In particular, by using variable discretisation it is possible to consider a much finer discretisation for variables $\gamma$ and $\delta$ than is used in the work by Jongerden et al. (2009) and to handle very sensitive interactions. This is crucial, particularly when the available charge in the batteries is almost exhausted. Jongerden et al. (2009) describe their plans as optimal, but it is important to note that this is only with respect to the discretisation that they use; a finer-grained discretisation offers the opportunity for a higher quality solution to be found at the cost of a much larger state space. Despite the very large state space our model creates, the solver visits a very small collection of states (as shown in the table). These problems are all solved in less than a second.

When dealing with larger batteries of type $B_2$, the state space becomes so large that any exhaustive approach is infeasible. Indeed, in the works by Jongerden et al. (2009, 2008), the authors were not able to handle this second case. We also found high quality solutions for batteries of type $B_2$: an example is shown in Figure 11 compared with the standard best-of-two solution, showing the huge improvement we can obtain over this policy. Note that the slicing of the load periods occurs towards the end of the plan, and this is a phenomenon we have observed in all our plans.

We also considered an 8 battery system (an example of its behaviour is shown in Figure 14). Benini et al. (2003) indicate that the designers of the SMBus (SBS Implementers Forum, 2000) architecture, which is a communication and control architecture and protocol that has been used in the development of Smart Batteries, suggest that there might be good reasons not to partition charge among more than four batteries. In fact, there are examples of systems using more than four batteries, such as HP 6-cell lithium-ion Smart Battery packs. In practice, partitioning charge between batteries offers multiple benefits, including the opportunities to use industry standard cells and to exploit different distributions of weight and possible cooling requirements. The tradeoffs between these benefits and the potential loss of efficiency arising from the partitioning is complex. The more batteries that are to be used, the larger is the state space for both planning and policy learning; constructing a solution to an 8 battery problem is significantly harder than for a 4 battery problem, so we present these results as evidence that we can scale to larger systems, subsuming the smaller cases.

The results are reported in Table 2, and show that we can scale effectively to much larger problems. Notice that the number of switches we use to produce the results is very significantly smaller than the best-of-8 policy giving the theoretical upper bound, however the resulting solutions achieve more than 99% efficiency. The final column, labelled *Plan-based Policy*, shows the performance of the policies we discuss in the next section, applied to these load profiles. These generate slightly worse performance in switches, but maintain the lifetime performance.

One final observation worth noting is that the structure of the usage profile across the batteries leads, in the two-battery case, to one battery being discharged sooner than the other. In the 8-battery case this effect is more pronounced, with several batteries being discharged while others still have significant charge remaining. This has an interesting consequence: using this policy it becomes possible to "hot-swap" batteries, replacing used batteries with new ones, while the system is active. The fact that one or more batteries still hold charge allows loads to be serviced while the used batteries are exchanged with charged ones and the policy can adapt to the new states of charge of

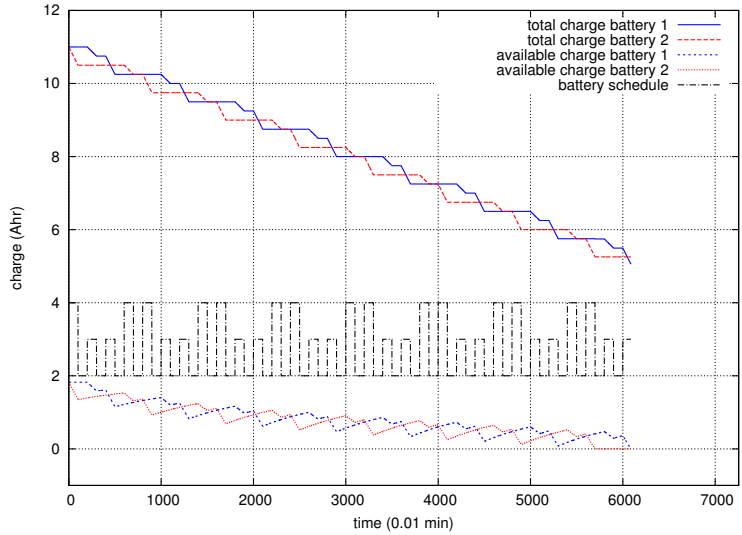

(a) $V_{max}$ (based on the feasible frequency switching used in (Jongerden et al. 2009))

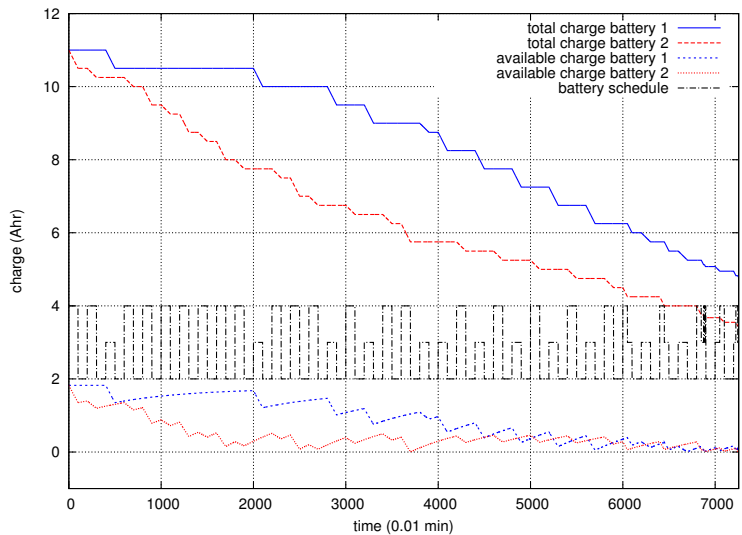

(b) Plan

Figure 11: ILs_alt load test with two batteries of type $B_2$

the batteries once the used ones have been replaced. This is in marked contrast to the high-frequency switching policies, where the batteries all discharge at approximately the same time.

## 5. From Plans to Policies

Having shown how to generate high quality plans for deterministic multiple battery management problems, we now turn our attention to the stochastic problem we are really interested in solving. In general, we cannot know in advance what will be the load profile applied to the batteries, but we assume that a probability distribution characterising typical use of the batteries is available. Such a probabilistic problem can be cast as a hybrid temporal Markov Decision Process (MDP).

| load profile | 8 batteries $B_2$ lifetime (number of switches) | | |
|---|---|---|---|
| | Upper bound | Plan | Plan-based Policy |
| CL_250 | 310.6 (31072) | **307.6 (485)** | **307.6 (992)** |
| CL_500 | 134.7 (13472) | **133.4 (266)** | **133.4 (571)** |
| CL_alt | 192.8 (19280) | **190.8 (355)** | **190.8 (806)** |
| ILs_250 | 660.7 (33076) | **654.1 (495)** | **654.1 (904)** |
| ILs_500 | 308.7 (15476) | **305.7 (293)** | **305.7 (513)** |
| ILs_alt | 424.8 (21280) | **420.6 (357)** | **420.6 (614)** |
| IL$l$_250 | 1008.9 (33692) | **998.8 (471)** | **998.8 (822)** |
| IL$l$_500 | 480.9 (16090) | **476.1 (295)** | **476.1 (597)** |

Table 2: System lifetime (in minutes) for all load profiles serviced with 8 batteries

Formally, a MDP is defined as follows:

**Definition 4** *A Markov Decision Process is a 4-tuple, $(S, A, P, R)$, where $S$ is a set of states, $A$ is a finite set of actions, $P$ is a probability function where $P_a(s, s') = Pr(s_{t+1} = s' | s_t = s, a_t = a)$ is the probability that action $a \in A$ will cause a transition from state $s \in S$ to $s' \in S$ when applied at time $t$, and $R$ is a reward function, where $R_a(s, s')$ is the reward earned for making the transition from state $s$ to $s'$ by action $a$.*

The Markov property is that the probability distribution for a transition out of a state is not affected by the path by which the state was reached. In general, MDPs are defined with finite state spaces, but a continuous MDP can also be considered, in which the states are embedded in multidimensional real space. The battery usage problem can be seen as a continuous MDP, where the states are tuples that define the (continuous) state parameters for each of the batteries and also the current state of the load and which battery is servicing the load (if the load is non-zero). Actions in this problem indicate which battery should now service the load, but can also correspond to events that change the current load. In the battery problem the actions switching between batteries are deterministic, but the events that cause load changes are probabilistic, representing the uncertainty about the demands of the user on the powered system. The time between events is also governed by a stochastic process, but the timing of switching actions is controllable.

More formally, for a problem with $n$ batteries, a state is characterised by the tuple $(sb_1, sa_1, sb_2, sa_2, ..., sb_n, sa_n, B, t, L)$, where $sb_i$ is the bound charge in battery $i$, $sa_i$ is the available charge in battery $i$, $B$ is the number of the battery currently servicing load ($1 \leq B \leq n$), $t$ is the time of the state and $L$ is the current load. Out of each state there is a deterministic action, *Use* $B'$, which causes a transition to the state $(sb_1, sa_1, sb_2, sa_2, ..., sb_n, sa_n, B', t, L)$, in which battery $B'$ is the battery servicing load. There is also a non-deterministic action, *wait(T)*, where $T$ is a time interval, which causes a transition to a state in which time has advanced to time $t' \leq t + T$, the state of charge of battery $B$ is updated according to the battery model and the load might be different (according to the probability distribution governing loads). The interpretation of the action is that it advances time to the next event, which will be when a battery is depleted of available charge, or when the load changes, or when $T$ time has passed, whichever is first.

The reward function for the battery problem gives positive reward for each transition, proportional to the advance of variable $t$. Once the system enters a state in which the currently active battery has no available charge, it terminates (or, equivalently, enters a special final state on which all further transitions loop without incrementing $t$). This reward system means that the optimal solution will be the one with greatest duration.

A solution to an MDP is a *policy*:

**Definition 5** *A policy, $\pi$, for MDP $(S, A, P, R)$, is a mapping $\pi : S \to A$, specifying which action to execute in each state.*

For the battery problem, the policy will be a function that determines which battery to use when load must be serviced, using the current states of charge of the available batteries as the basis for making the decision.

Considerable research effort has been invested in the problem of finding policies for MDPs, as discussed in Section 6.

The way we approach this problem is to see the mapping as a classification, where the state of the batteries is mapped to a class corresponding to the correct choice of battery. We can use the solutions to the determinised problems as the basis of a classifier construction problem and use an existing machine learning approach to build a good classifier. The overall approach is sketched in Figure 12.

Several important observations can be made. Firstly, the successful construction of a classifier depends on there being exploitable structure in the space defined by the solutions to the determinised problems. Secondly, the states are described by continuous variables: we discretise these for the purpose of building the classifier. Thirdly, our solution set will generally not cover the whole space of reachable states, so it is important that we complete the policy with a sensible *default action* to deal with states that the policy fails to handle. In our case, the default action is a best-of-$n$ rule, which is the best of the published hand-constructed policies for this problem. If the policy suggests to switch to a battery whose available charge is below a critical threshold, then the policy action is ignored, and the default action is used. We discuss the impact of this in physical experiments in Section 7.

Finally, we note that deployment of the constructed policies will require that they can be efficiently implemented in cheap hardware. Simple classifier rule systems can be very effectively implemented in look-up tables, which are ideal for implementation on Field Programmable Gate Arrays (FPGAs) or as purpose-built hardware.

## 5.1 Policy Learning through Classification

To learn a policy through classification, it is first necessary to generate an appropriate training data set. For our problem, this data set must associate the states of the batteries and the current load with an appropriate decision (which battery to use to service the load). We construct the training set by building a sample of profiles from the stochastic description of the expected loads. The distributions we used to describe amplitude, duration and frequency of loads are shown in Figure 13. The deterministic solutions to these problems are constructed as described in Section 4. Training data is then generated from these plans by simulating their execution and recording the battery states, load and battery choice at a fixed time increment throughout the plan. For example, if the increment is 0.01 minutes then the training data generated from a plan will record the battery states of charge (available and bound), load and currently selected battery (which might or might not have changed from the previous time increment) at every 0.01 minute interval throughout the plan. In our experiments we selected the time increment to be the same as the smallest increment used in the variable discretisation described in Section 4.4, but this is not a requirement of the approach. The choice of time increment determines the frequency of the decision-cycle for the learned policy. The time increment also determines how much training data is generated from a single plan, according to the makespan of the plan. In order to reduce the volume of training data for fine-grained time increments used with long makespan batteries, it is possible to randomly sample from the set of state-battery-selection pairs across multiple plans. In our experiments we did not need to do this.

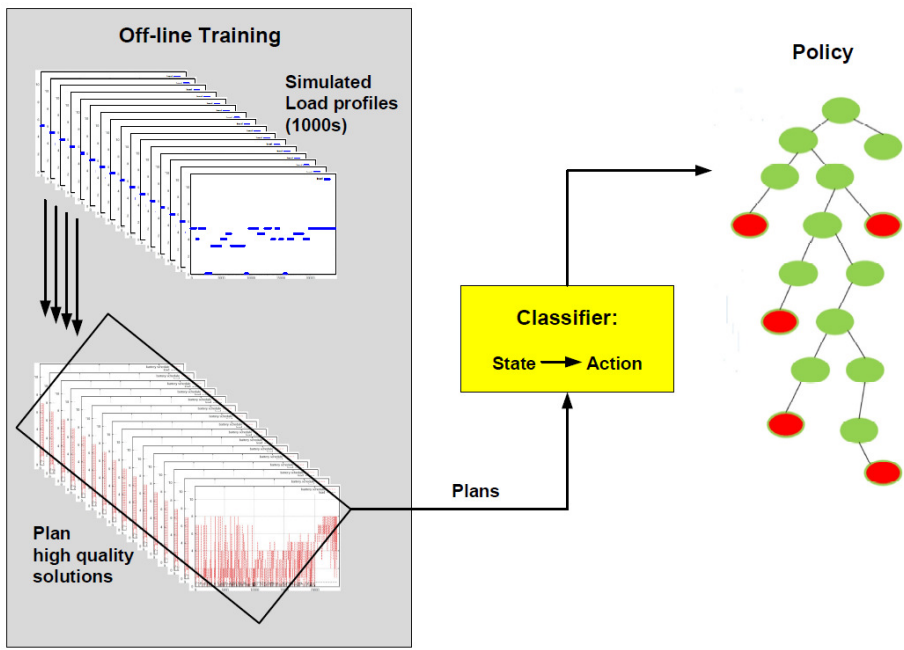

Figure 12: Plan-based policy learning: this figure illustrates our approach to policy learning schematically. The off-line training phase involves construction of a set of planning problem instances by sampling from the initial state distribution, followed by construction of plans for each instance. These plans are then classified to obtain a state-to-action mapping in the form of a decision tree which can be used as a policy.

Once the training data is generated, a classifier can be learned using a standard machine learning approach. WEKA (Hall, Frank, Holmes, Pfahringer, Reutemann, & Witten, 2009) is a machine learning framework, developed at the University of Waikato, that provides a set of classification and clustering algorithms for data-mining tasks. WEKA takes as input a training set, comprising a list of instances sharing a set of attributes. In order to perform the classification on the battery usage problem data, we consider instances of the following form:

$$\tau = (\sigma_1, \gamma_1, \ldots, \sigma_N, \gamma_N, B, L)$$

where $\sigma_i$ and $\gamma_i$ denote the available charge and total charge of the $i$th battery, respectively, $B$ is the currently active battery and $L$ is the current load (this is essentially the state of the MDP but without the time label, since we want our policy to operate independently of time). In this setting, the attribute used as the class is the battery $B$.

The stochastic load profiles have been defined with a distribution of:

- the load amplitude $l \in [100\dots750]$ mA;

- the load/idle period duration $d \in [0.1\dots5]$ min;

- the load frequency $f \in [0.3\dots0.7]$.

The probability distributions are shown in Figure 13.

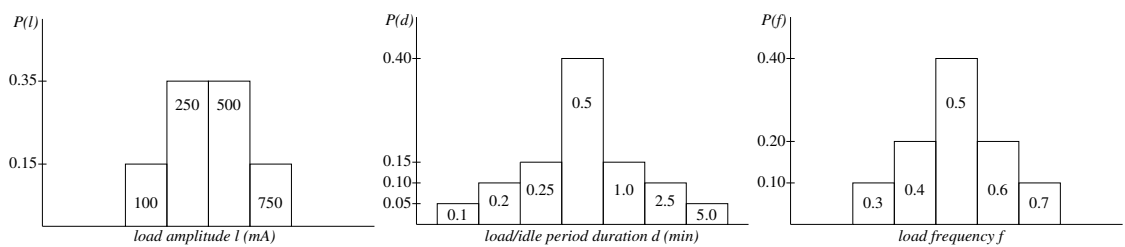

Figure 13: Probability distributions for the stochastic load profiles

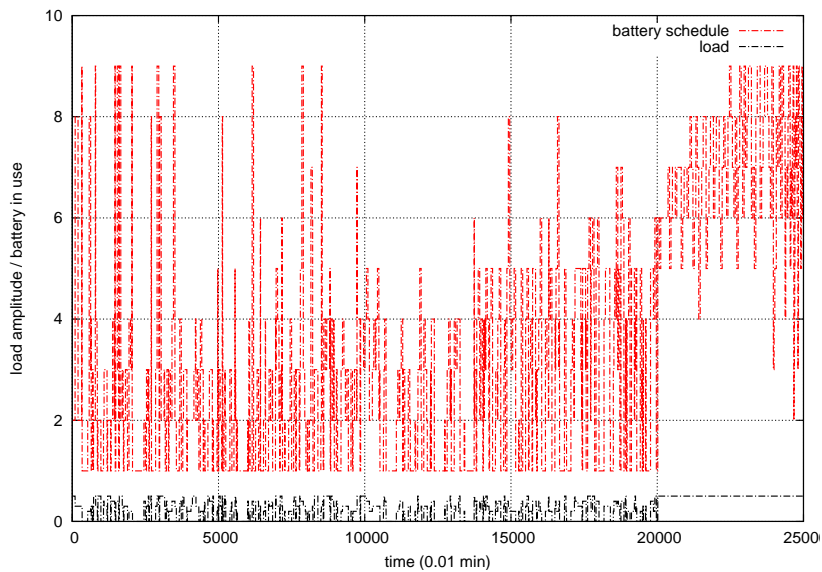

Figure 14: Plan-based policy for 8 batteries with a stochastic load

This leads to load profiles that are very irregular (see the bottom of Figure 14) and therefore harder to handle than the very regular profiles considered by Jongerden et al. We generated a set of stochastic load profiles and for each of them we produced a near-optimal plan using the deterministic solving described in Section 4. This set of plans has been used as the training set for the classification process.

| Algorithm | cross-validation success | model size |
|---|---|---|
| DMNBtext | 18% | – |
| NaiveBayes | 37% | – |
| NaiveBayesSimple | 36% | – |
| NaiveBayesUpdateable | 36% | – |
| Logistic | 44% | – |
| MultilayerPerceptron | 51% | – |
| RBFNetwork | 43% | – |
| SimpleLogistic | 44% | – |
| SMO | 44% | – |
| IB1 | 99% | 26 Mb |
| IBk | 99% | 26 Mb |
| AdaBoostM1 | 27% | – |
| AttributeSelectedClassifier | 98% | 29 Mb |
| Bagging | 98% | 18 Mb |
| Clustering | 26% | – |
| Regression | 98% | 9 Mb |
| CVParameterSelection | 19% | – |
| Dagging | 44% | – |
| Decorate | 99% | 31 Mb |
| END | 99% | 15 Mb |
| EnsembleSelection | 99% | 70 Mb |
| Grading | 19% | – |
| LogitBooost | 47% | – |
| RandomCommittee | 99% | 12 Mb |
| RandomSubSpace | 99% | 21 Mb |
| RotationForest | 99% | 22 Mb |
| Stacking | 19% | – |
| Vote | 19% | – |
| VFI | 23% | – |
| DecisionTable | 90% | 6 Mb |
| DTNB | 90% | 6 Mb |
| DecisionStump | 27% | – |
| **J48** | **99%** | **2 Mb** |
| J48graft | 99% | 13 Mb |
| OneR | 56% | – |
| LADTree | 45% | – |
| NBTree | 99% | 114 Mb |
| SimpleCart | 99% | 86 Mb |

Table 3: Performance of classification algorithms tested on 10,000 training examples

In order to select the most suitable classification algorithm, we applied all the classifiers provided by WEKA to a data set of 10,000 training examples. We first evaluated their *performance* as the number of correctly classified instances during the cross-validation. We discarded classifiers providing less than 70% correctness. We then considered the *memory* and the *time* required to use the classifier. The output of the classification process is a model encoding the resulting decision tree. In some cases, the generated model requires significant memory to store (more than 500Mb of RAM memory), or it is too slow to be used. These parameters have also been used to determine the number of training examples to classify, as the bigger the training set, the better the performance and the higher the memory and time requirements. Some of the classifiers with their performance are reported in Table 3.

```
...
if(b2gamma<=0.297404){
    if(b2gamma<=0.296404){
        if(b2gamma<=0.288404){
            if(b2gamma<=0.286404){
                if(b2gamma<=0.277404){
                    return 1;
                }
                if(b2gamma>0.277404){
                    return 2;
                }
            }
            if(b2gamma>0.286404){
                return 1;
            }
        }
        if(b2gamma>0.288404){
            return 2;
        }
    }
    if(b2gamma>0.296404){
        if(b2y1<=-0.043615){
            return 1;
        }
        if(b2y1>-0.043615){
            if(b1gamma<=0.164404){
                return 1;
            }
            if(b1gamma>0.164404){
                return 2;
            }
        }
    }
...
```

Figure 15: Fragment of decision tree

According to these criteria, we selected the `J48` classifier, which implements the machine learning algorithm C4.5 (Quinlan, 1993). The output is a decision tree whose leaves represent, in our case study, the battery to be used (a fragment of the tree is shown in Figure 15). For the cardinality of the training set, an empirical evaluation showed that the best result is obtained using 250,000 training examples (note that this involves considering about $4 \cdot 10^6$ real values characterising the states and battery selections in these training examples) since further extending the training set does not make any significant improvement in the performance but increases memory and time requirements.

| load | Upper bound | | Plan-based Policy | |
|---|---|---|---|---|
| profile | time$_{(\sigma)}$ | $sw_{(\sigma)}$ | time$_{(\sigma)}$ | $sw_{(\sigma)}$ |
| R100 | $792.6_{(15.5)}$ | $71383_{(1379)}$ | $\mathbf{786.2}_{(15.4)}$ | $\mathbf{1667}_{(161)}$ |
| R250 | $369.8_{(1.91)}$ | $28952_{(853)}$ | $\mathbf{366.7}_{(2.02)}$ | $\mathbf{1518}_{(143)}$ |
| R500 | $226.7_{(2.13)}$ | $14671_{(512)}$ | $\mathbf{224.6}_{(2.27)}$ | $\mathbf{987}_{(122)}$ |
| R750 | $188.3_{(0.8)}$ | $11519_{(463)}$ | $\mathbf{186.4}_{(0.7)}$ | $\mathbf{302}_{(33)}$ |

Table 4: Average system lifetime and number of switches for stochastic load profiles for 8 battery systems

## 5.2 Results from Policies

In order to use the decision tree we embedded the WEKA classes for loading the classification model into our battery simulation framework. The model for the 8 battery case is represented by a tree with 61 levels and consists of 7645 nodes, each one containing a comparison between one of the state variables and a threshold. Applying this decision tree to determine which battery to load at each decision point takes negligible time.

To evaluate the performance of the policy we considered four probability distributions with different average value for the load amplitude, namely 100, 250, 500, 750 mA. For each distribution we generated 100 stochastic load profiles and we used the policy to service them. Note that the load profiles used for evaluating the policy are independent from the ones used for training, although they are drawn from the same probability distributions.

Table 4 shows the average value and standard deviation for the system lifetime and the number of switches obtained using the best-of-8 policy at high frequency switching and our policy.

Also in this case, we observe that our policy achieves more than 99% efficiency compared with the theoretical upper bound given by the best-of-8 policy executed at very high frequency (recall that this is infeasible in practice). Moreover, the number of switches used by the policy is slightly greater than in the corresponding deterministic solving, but is one order of magnitude lower than the corresponding value for the best-of-$n$ policy.

## 6. Related Work

A variety of approaches have been proposed for solving continuous Markov Decision Processes (Sanner & Boutilier, 2009). Meuleau et al. (2009) propose hybrid AO* search, using a dynamic programming approach to guide heuristic search for problems involving continuous resources used by stochastic actions. This approach does not handle time-dependent resource consumption, but it appears that the above MDP could be modelled for solution by this approach. The authors give empirical data for solution of problems with up to 25,000 states. Our model, with an appropriate discretisation, contains more that $10^{86}$ states for 8 batteries. Mausam and Weld (2008) describe a planner for concurrent MDPs, which are MDPs with temporal uncertainty. Again, these problems are similar to ours, although their planner does not manage continuous time-dependent resources, so is not directly applicable to our problem. Furthermore, the largest problems they consider contain 4,000,000 states and take more than an hour to solve.

In solving very large MDPs, researchers have identified a variety of techniques that can help to overcome the prohibitive cost of policy iteration or value iteration, the classical techniques for solving MDPs. In general, these techniques approximate the solution, often focussing on those parts of the policy that apply to states that are likely to be visited along the trajectory. Relevant techniques are discussed in the work of Bertsekas and Tsitsiklis (1996).

Our approach is in the branch of work devoted to the development of plan-based reasoning under uncertainty. In fact, when explicit modelling of uncertainty is impractical, sampling can provide an effective alternative.

*Hindsight Optimisation* (HO) (Chang, Givan, & Chong, 2000; Fern, Yoon, & Givan, 2006) has become a well-researched technique for learning policies based on plans. A policy always proposes the best action to do next in any state, and is therefore more or less robust to the uncertainty encountered in reality. The HO technique works as follows: given an MDP and a state, $s$, the first step is to sample, from the MDP, a large number of deterministic instances of the process with initial state $s$. The next step is to solve these instances using a deterministic planner over a fixed horizon. Finally, the estimated value for the state $s$ is computed as the *average* value obtained from the deterministic plans. It is then possible to choose, in any state, the move that led to the best performance on average in the samples.

Although our approach is similar to Hindsight Optimisation, there are significant differences.

First, previous works in this direction have only addressed propositional domains (see, e.g. the work of Fern, Yoon and Givan (2004, 2006, 2007), or Königsbuch, Kuter and Infantes, 2010) while here we are interested in a *hybrid discrete-continuous problem*, as we deal with the non-linear continuous and deterministic planning models of the drain and recovery behaviour of batteries, using sampling to provide the noise encountered in reality. The approach is to sample the deterministic instances of the problem using simple assumptions about the underlying distributions governing the physical reality. In many natural situations, Gaussian distributions work well as an approximation of the uncertainty in the problem. In this work, for example, we show that by sampling many deterministic discretised cases, and planning solutions to each of them exactly, it is possible to classify the states of the solution plans into a policy that can *robustly* manage the load distribution in both simulated and real battery configurations. The weaknesses of the assumptions made about the underlying distributions are overcome by introducing default actions (described in Section 5), which can be applied when the policy finds itself in a state outside the range of applicability of the policy. Integrating the policy with the default action leads to very competent policies that perform well across a wide range of physical situations, including situations that are dissimilar to those encountered during the learning phase.

Another important difference is that rather than averaging over plan states to obtain a policy, in our approach we use a *decision tree classifier* to arrange the states according to their information content (reflected in how well they support a partitioning of the planned actions). This results in a classification of actions into states, and a policy that proposes the best action to use in any state is determined online by comparing policy state variables with the real values encountered as the policy is executed. Although training for policy-learning is expensive in terms of time and computational resources, the planning and learning is done offline, and the offline process is not strongly resource-bounded. The classification phase produces a policy in the form of a decision tree, that is compact and the execution of which takes negligible time and this is a key feature for this application. In fact, due to the continuity involved in the battery model, and the need for planning to a very long horizon (up to 60,000 time steps), the resulting state space is huge. This makes any approach based on an explicit mapping of each state to an action impractical. In particular, it is not possible to compute an HO-based policy offline and then map each state to the best action according to the policy values. On the other hand, using HO online (which is viable in many cases) in infeasible in this application, as the nature of the battery scheduling problem requires a very fast interaction between the policy and the battery system. Our approach meets both the *scalability* and *fast-response* requirements.

Finally, the idea of looking ahead over "what if" scenarios, and then benefiting from the experience gained, is powerful. In HO it is assumed that, in general, the experience of the deterministic planner is sufficient to give insights into the best moves possible in a real state encountered during execution. However, another important aspect that makes our approach different, and that we investigated more deeply in a different context (Fox, Long, & Magazzeni, 2012), is that, in many cases, it is necessary to distinguish between the *plan state* and the *policy state*. For example, while the plan state might contain a variable representing whether an unreliable valve is open or closed, observable experience records the effects of its unreliability – for example, the effect on flow-rate through a pipe – over a given time period. A policy-state variable can therefore be constructed to record the observed flow rate, which is a proxy for whether the valve is open or closed. This approach, which we call *observable-correlate policy learning*, is very different from averaging over the plan states encountered during planning, because policy states capture the actual situation being experienced, while plan states remain abstracted and distanced from reality. In that work (Fox et al., 2012), we apply exactly the same policy-learning technique as described here to the problem of learning robust observable-correlate policies for following the boundary of a surface algal bloom. In this context we define a collection of policy state variables which correlate plan state variables with observable experience.

## 7. Physical Experiments

In this section we report the results obtained from a 'kitchen table' experiment comprising a simple circuit constructed out of breadboard components and an Arduino Mega board which we used for sensing and control.[1] Using this apparatus we have been able to demonstrate that our simulation results do translate into reality. As part of our future work, further experiments will be undertaken in a professional laboratory to continue to explore the benefits and limitations of our approach.

The goal of the experiment is to demonstrate that the plan-based policy method achieves similar lifetime to that achieved by the best-of-two policy, but with significantly reduced switching. It is clear from the simulation results that the plan-based policy can achieve close to optimal lifetime with only a fraction of the switching that best-of-two requires, although the simulation also suggests that the best-of-two policy should achieve within less than 1% of the theoretical optimal even switching at a frequency of once every 5 minutes. We therefore expected little opportunity for our learned policy to improve the lifetime and were therefore hoping to achieve similar lifetime but with a much lower switching frequency. Our results show that the plan-based policy does exhibit much lower frequency switching. In fact we found that the plan-based policy achieves significantly longer lifetimes as well.

---

1. The results and figures presented throughout this section are presented in colour in order to clarify the relationships between multiple plots. Unfortunately, several figures are difficult to interpret in monochrome and the reader is recommended to view the figures using an appropriate medium.

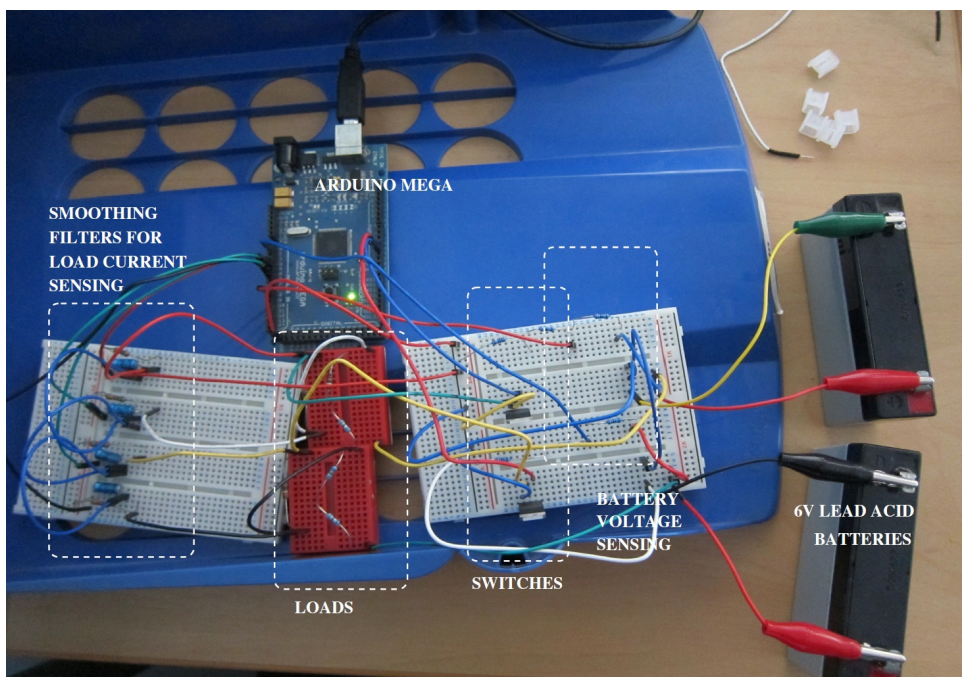

Figure 16: A photograph of the battery apparatus constructed to manage two batteries.

We begin by describing how we built the circuit that we used for the experiment. We then recall the KiBaM model, and explain how its parameters were estimated. The plan-based and best-of-two policies rely on being able to read the state of available charge of the batteries. This is very difficult to estimate, and the performance of the policies depends absolutely on estimating this quantity accurately, so we explain how we read state of available charge in our set-up. Finally we present the results of our experiments and describe our plans for future work.

## 7.1 The Electronic Apparatus

We constructed an experimental apparatus for a suite of two batteries, shown in Figure 16. We used Ritar 6 volt lead acid batteries of nominal capacity 1 Amp hour for 20 hours of discharge (1Ah@20h). We connected each of these batteries in a circuit to an Arduino Mega board.

Part of each circuit was constructed to allow the Arduino to read the voltage on the connected battery. We want to ensure that the current drawn to measure the voltage is negligible, so high external resistance, of 3.6k$\Omega$ and 7.2k$\Omega$, was used to bridge the Arduino input. Using a voltmeter we read 6.5-6.7V on a fresh battery, so we consider V$_{EMF}$ = 6.5V. This is too high a voltage for the Arduino inputs which have a maximum input voltage of 5V. Since, considering the battery voltage sensing element of the circuit with resistance $R$, $V_{EMF} = iR$ and $V_{EMF} = 6.5V$, we use $R = 7.2 + 3.6 = 10.8k\Omega$ in order to divide the voltage and to achieve a negligible current of 0.0006A. A higher resistance might seem preferable to still further reduce the current losses, but the

Figure 17: The battery apparatus for two batteries.

Arduino uses an analog-to-digital converter based on measuring charge on a capacitor over time. This approach relies on sufficient current flow into the capacitor to get accurate measurements in short time periods and very high resistance prevents this. In practice, a resistance of $\sim 10k\Omega$ is about the limit at which the Arduino can respond to changes in the inputs within the timing constraints of our sampling. With these resistances the voltage reading at the Arduino is $V_{EMF} - 0.0006 \times 3600\Omega = 4.34V$, which is within its operating range.

The current is diverted to a load consisting of a switch and two resistors of 8 and $1\Omega$. The role of the switch, which is a MOSFET IRF630 controlled using a pulse width modulated output from the Arduino, is to ensure a smooth delivery of power to the resistors. The load is $6.5/(9+r+R_s)$ where $r$ is the internal resistance of the battery and $R_s$ is the effective variable switch resistance under pulse width modulated control. The data sheet for the Ritar 6V battery lists the internal resistance, $r$, as $50m\Omega$, while we measured $0.34\Omega$, a value almost 7 times greater. We believe that the discrepancy comes from a systematic distortion in the sensed values reported by the Arduino. We consistently use these readings in all of our experiments and regard the discrepancy as a systematic error. Our experiments use currents varying between $0.2A$ and $0.3A$, so, when $V_{EMF} = 6.5V$ and $i = 0.3A$, $R_s$ is about $12\Omega$, but is lower when the battery is less charged (and the voltage drops) and higher when a lower current load is required.

The circuit diagram is shown in Figure 17. It will be noted that the load is duplicated in this design, which completely separates the parts of the circuit responsible for interacting with each battery. In fielded systems the load would be common and diodes used to prevent flow of electricity between batteries at different charge states.

## 7.2 Estimating Parameters

In this work we used the Kinetic Battery Model (Manwell & McGowan, 1993) and we followed the parameter estimation process described by Manwell and McGowan (1994). Following their description, the extended KiBaM has three parts: a capacity model, a voltage model and a lifetime model. We use a simple lifetime model (we assume that there is no change in the battery behaviour due to recharging).

### 7.2.1 THE CAPACITY MODEL

The capacity model, which describes how capacity varies as the battery is drained and allowed to rest, is described by a first order differential system. The quantity

$$q_{max}(I)$$

is the maximum amount of charge, in Amp hours, that we could hope to extract from the battery if we discharged it continuously, at nominal current $I$, until drained. The time it takes to drain the battery at nominal current $I$ is $T$. $T$ and $I$ are linked by the following equation:

$$q_{max}(I) = \frac{Ck'cT}{1 - e^{-k'T} + c(k'T - 1 + e^{-k'T})}$$

derived from the model described in Section 3.2. The model relies on three constants: $C$, which is the maximum capacity of the battery in Amp hours, $k$, which is the rate per hour of conductance between the bound well and the available well of the model, and $c$, which is the ratio of available charge to maximum capacity. In Section 3.2, $k'$ is defined to be $\frac{k}{c(1-c)}$. It can be seen that $q_{max}(I) = IT$.

These constants are found by fitting a curve to data. We obtained our data by draining batteries one at a time, from their fully charged state, using different currents in the circuit described in Section 7.1. An example of the data collected is shown in Figure 18, where the top curve is the measured voltage of the battery over time, the line at $5.25V$ is the point at which the battery is considered dead, the point cloud comprising a thick curve at $208mA$ is the measured load, and the thin straight line running through this point cloud is a rolling average of the load. The vertical line shows where we treated the battery as dead. As shown in Figure 19, there is uncertainty about exactly where the battery dies.

The values of $C, k, c$ that we calculated are:

$$C = 1.372Ah$$

$$k = 0.1967h^{-1}$$

$$c = 0.3870$$

and

$$k' = 0.8290h^{-1}$$

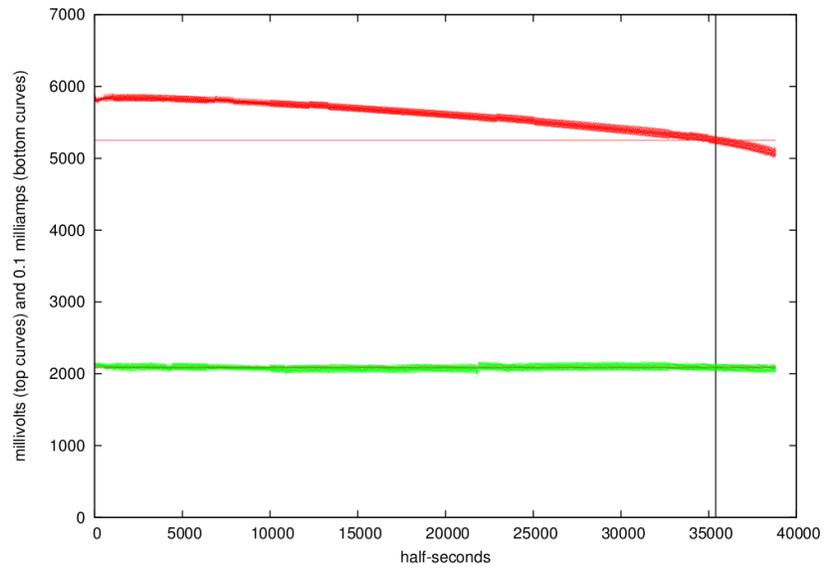

Figure 18: Battery discharge curve: terminal voltage (top curve) and load (bottom curve).

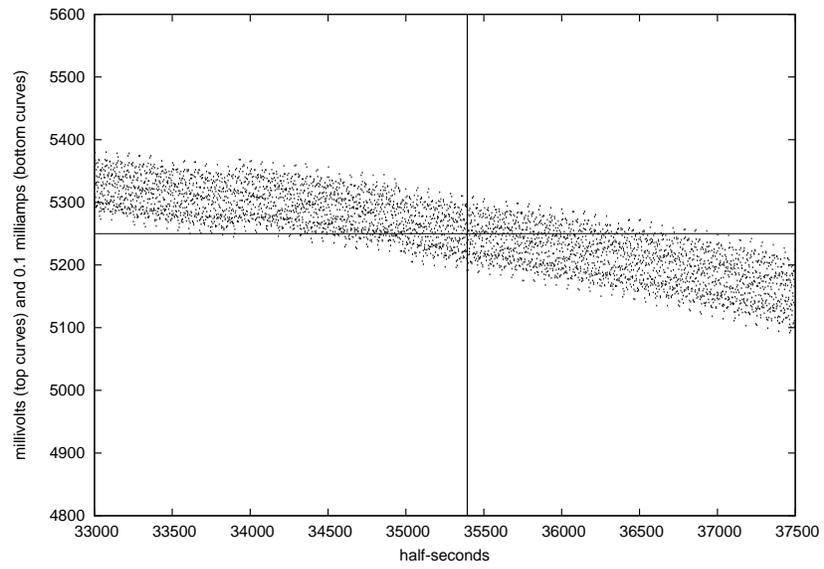

Figure 19: A close up of the point cloud of the voltage curve at the point where the battery is considered dead.

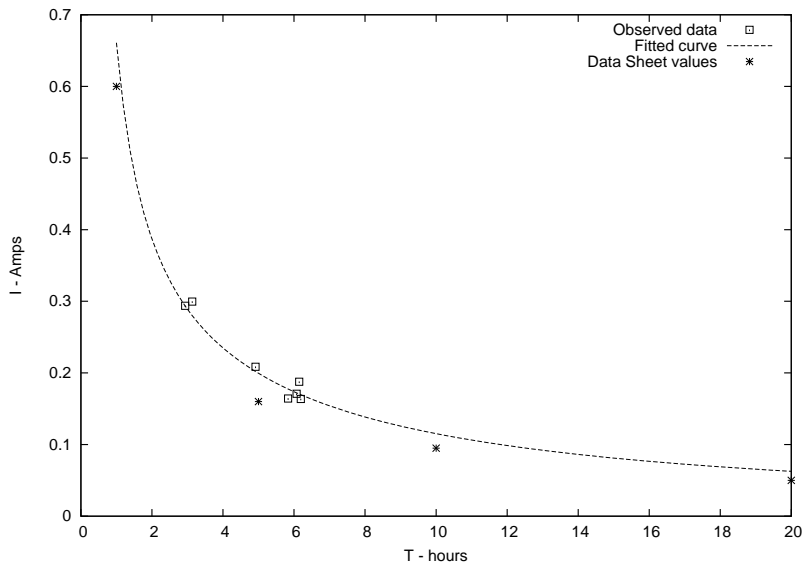

Figure 20: Data for current and time to drain batteries. The Data Sheet values are shown for comparison.

The fitted curve of $T$ against $I$, for the fitted $C, k, c$ values, is shown in Figure 20. The square points are our observed data, while the stars are the data points reported on the Ritar 6V battery data sheet. We found that the data sheet appears to consistently under-estimate the performance of the battery. It can be seen that our observed data points are clustered in the $0.17A$ to $0.3A$ region of the curve. We were unable to report points for lower currents, because the pulse width modulation could not be set to an appropriately low value without dropping the control voltage for the MOSFET switch below the point at which the switch opens. We could not report points for high currents without melting the resistors comprising the load on the circuit.

We used the $C$, $k$, and $c$ values to construct the initial state of the battery load management planning problem, and then we learned a policy from plans produced against this model. Therefore, an accurate estimation of these parameters is very important. The policy will be far less effective if the wrong capacity model is used. We learned a policy using a time granularity of 0.01h, which is 36 seconds. In our timing loops for collecting data from the Arduino sensors we use averages computed over 0.5 seconds: the data points in Figure 19 are shown at this resolution. Thus, we collect 72 data points from each sensor between decision points at the granularity of our planning model and, consequently, our learned policy. As can be seen, there is considerable noise in these values and to reduce this noise we construct a rolling average over the preceding window of 65 points. We selected 65 to avoid the particularly noisy data values generated when there is a switch between batteries.

### 7.2.2 THE VOLTAGE MODEL

In order to be able to exploit our plan-based policies it is necessary to be able to evaluate the state of charge of the batteries at every decision point. It is known to be very difficult to accurately evaluate state of charge because the behaviour of batteries is noisy, variable and highly non-linear. However, terminal voltage is recognised as a reasonable proxy for state of charge. We therefore observe the output voltage of each battery and calculate its state of charge from this reading.

The measured terminal voltage, $E_{obs}$, falls off as the battery is drained, producing a typical "knee-shaped" curve representing the decrease in voltage over time as the current is drawn, and illustrating the collapse in voltage once the battery is dead. Manwell and McGowan model this voltage curve using the equation:

$$V_{obs} = V_{EMF} + AX + BX/(D - X)$$

where $X$ is defined to be $\frac{Q}{q_{max}(I)}$ and $Q$ is the total charge consumed to date by the battery.

The parameters $A$, $B$ and $D$ are found by non-linear curve fitting to data, using voltage against time for constant current discharges. We used 4 sets of data obtained by draining batteries from fully charged, one at a time on our battery apparatus, to estimate the curve for the Ritar 6V batteries. Figure 21 shows an example of a discharge curve. The batteries are effectively dead as soon as the voltage drops over the knee. This occurs at $5.25V$. Figure 21 also shows a voltage model curve (the solid black line), of the type described above, fitted to the discharge data for a battery. In this case we have discharged the battery past the critical point where it is considered dead, to show how the voltage drops dramatically (and the load cannot be maintained reliably). The vertical line shows the point at which the battery is judged dead and the curve is fitted to the data up to this point. As can be seen, the curve fits well until after the knee, when the behaviour is no longer governed by the simple quadratic voltage model.

The parameter values we computed for our batteries are:

$$A = -0.194 mVs^{-1}$$

$$B = -2.22 \times 10^{-3} mVs^{-1}$$

and

$$D = 1.05h.$$

$A$ governs the almost linear decay in voltage over the first part of the discharge curve and it is the easiest parameter to estimate accurately. $B$ and $D$ together determine the shape and initiation of the dip in the voltage as the battery gets close to its dying threshold. The fit of the values for $B$ and $D$ is much more sensitive to noise than is the value of $A$.

### 7.2.3 EVALUATING THE STATE OF CHARGE OF THE BATTERY

Using the Arduino Mega board, we collect voltage and current values from the batteries at a frequency of every half a second. For each battery in use, we compute a rolling average over the last 65 voltage readings reported since the battery was first loaded (before this, the reported voltage readings can be inaccurate). Having computed the first rolling average we can fix $V_{EMF}$, which is the value we take to be the fully charged open circuit voltage of the battery (ie: the voltage that was available before any load was serviced). We calculate $E_{obs}$ and $Q$ every 36 seconds for every battery.

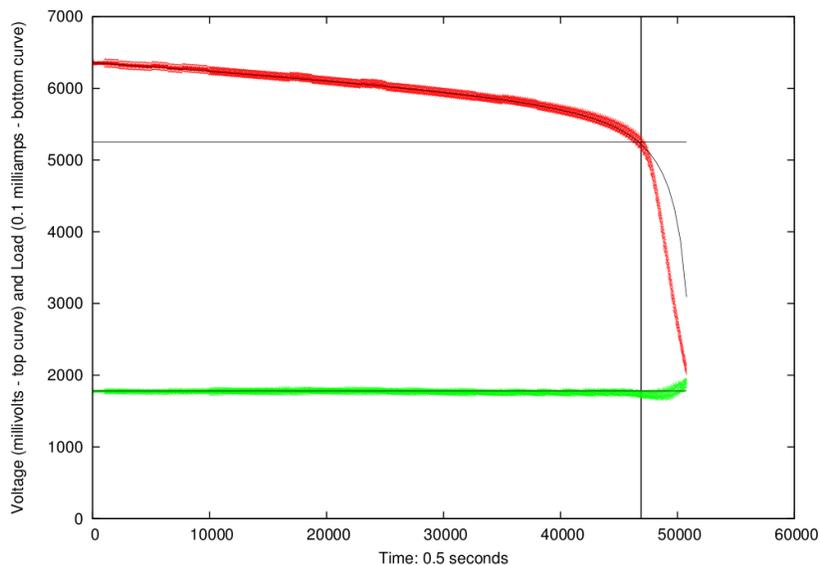

Figure 21: Voltage against time.

The observed voltage is affected by the load on the battery at the time that we observe it, so we adjust the observed voltage reading, $E_{obs}$, to take into account the internal resistance and load on the battery. This results in the unloaded observed voltage $V_{obs}$:

$$V_{obs} = E_{obs} + 0.34 I_{obs}$$

We can then calculate the difference between $V_{obs}$ and $V_{EMF}$ to be:

$$V_{adj} = V_{obs} - V_{EMF}.$$

Then, to calculate $X$ we first obtain a value $F$:

$$F = \frac{B + AD + V_{adj}}{2A}$$

Then:

$$X = F - \sqrt{F^2 - \frac{DV_{adj}}{A}}$$

We use this root of the quadratic equation for $X$ because $X \leq 1$.

For a given battery, $b$, to calculate the charge consumed by $b$ at time $t$, the sum of the current readings taken so far (measured in milliamps, taken every half second) is divided by a large constant, $7.2 \times 10^6$, which gives a result in Amp hours. This value is $Q$, the total charge consumed to date by $b$.

The value $X$, which is the proportion of available charge at current $I$ that has been drawn, is obtained from the two parameters $E_{obs}$ and $Q$, using the voltage model given above. Once we have $X$ and $Q$, we can compute $q_{max}(I)$ as $\frac{Q}{X}$.

We can now evaluate the state of charge of a battery. The variable $\gamma$ is the total capacity, $C$, minus the total charge consumed, $Q$, in Amp hours. This gives us an estimate of the total remaining charge, but not all of this will be accessible because some of it is bound up in the chemical properties of the battery. The variable $\delta$ is the difference between the bound and available charge wells, enabling us to estimate how long we would need to drain the battery. Since available charge will always be less than or equal to the bound charge, there is always a pair of values $(I_{nom}, T_{nom})$, such that, had the battery been run at $I_{nom}$ for time $T_{nom}$, it would have reached its current state of charge. Given that

$$X = \frac{Q}{q_{max}(I_{nom})}$$

and using the equation for $q_{max}(I)$ given in Section 7.2.1, we have that

$$\frac{Ck'cXT_{nom}}{Q} = 1 - e^{-k'T_{nom}} + c(k'T_{nom} - 1 + e^{-k'T_{nom}})$$

Therefore, $T_{nom}$ is the solution of

$$1 + (c - 1)e^{-k'T} + ck'(1 - \frac{CX}{Q})T = 0$$

The time, $T_{nom}$, that is nominally required to continuously drain the battery from fully charged, at current $I$, is calculated numerically by plugging these equations into the Newton-Raphson method, with an appropriate initial value (we use 4, since the expected lifetime of the battery at the discharge rates we are using is about 2-4 hours). Given that:

$$q_{max}(I) = I_{nom} \times T_{nom}$$

we have that:

$$I_{nom} = \frac{q_{max}(I)}{T_{nom}}$$

and $\delta$ is then computed as:

$$\frac{I_{nom}(1 - e^{k'T_{nom}})}{ck'}$$

The available charge can be calculated from $\gamma$ and $\delta$ as:

$$c(\gamma - (1 - c)\delta)$$

as discussed in Section 3.2.

The best-of-two policy discussed in Section 3 can now be implemented to always choose the battery with the highest available charge. Executing this policy requires the state of charge to be read with reasonable accuracy at the fixed frequency. For example, one might fix the frequency to be every 6 minutes, and select for the next 6-minute interval the battery with the highest available charge (which is equal to $c(\gamma - (1 - c)\delta)$ as explained in Section 3.2).

### 7.2.4 Recharging and Other Effects

It is clear that to perform multiple experiments with lead-acid batteries it will be necessary to recharge them between discharges. Recharging lead-acid batteries is known to have an impact on their performance: they deteriorate with repeated cycling. However, the gel-type batteries we used are deep cycle batteries that can be cycled hundreds of times before they reach the end of their design life.

Manwell and McGowan (1994) have proposed a lifetime model based on a rainflow cycle-counting algorithm which takes into account the fact that recharging damages the batteries and affects their ability to deliver charge. Given that our batteries were brand new, and we have used each one no more than 30 times, we hypothesise that the effects of repeated discharging and recharging will not be significant in the lifetime of our experiment[2]. For an extended, or larger scale experiment, the rainflow model would be of interest, but adopting it, and exploring how it changes the behaviour of our model, is left for future work.

An additional important effect on battery behaviour is temperature. All of our experiments were conducted in an office environment with normal working temperatures. One of the factors that governed our choice of discharge currents was the fact that at high discharge currents the batteries do warm up noticeably, so the model we are using is likely to cease to be valid without changes to the parameters. We ignored temperature effects and treat the batteries as though they are used at a constant standard operating temperature, which is a reasonable approximation.

## 7.3 The Experiments

We carried out three sets of experiments on an apparatus consisting of two Ritar 6V batteries connected to the circuit shown in Figures 16 and 17. In our simulation tests we demonstrated the performance of our approach on suites of 8 batteries, but performing the same experiments on the physical apparatus would have been too time-consuming. Each of our 2-battery experiments took over 11 hours to drain the batteries and, if anything went wrong during an experiment, such as loss of communications with the PC, the experiment had to be restarted resulting in the loss of a day or more.

When performing the experiments we noticed that the Arduino distorts all measured values: time and voltages, and therefore amps and internal resistance. Its distortions appear consistent across all experiments, resulting in systematic error. In particular, all of the times we measured suggest that the Arduino measures 1 hour every 1.4 hours of real time, so a 7 or 8 hour lifetime measured by the Arduino is actually approximately 10 to 11 hours of real time. We report all data values directly from the Arduino measurements, unadjusted for the systematic errors, so it can be borne in mind that our lifetime values are considerably longer when measured in "real" time. For consistency, all other times are reported in the same relative measures (in practice, timing of load control and discharge curves and other values were all performed using the Arduino clock, so the measurements are entirely consistent with one another).

We randomly generated 10 different load profiles, drawn from the same distribution as we used to train our policy, each alternating between 0.2 and 0.3 Amps and having intervals of constant load of durations that are distributed around 30 minutes with a distribution as shown in Figure 22.

---

2. The experiments we report for load profiles 1–6 were run with batteries having been cycled up to 15 times. For later profiles we did observe that some of the batteries showed behaviour that suggested a slight deterioration in performance and it is possible that lifetimes are lower for these experiments than would be the case for new batteries.

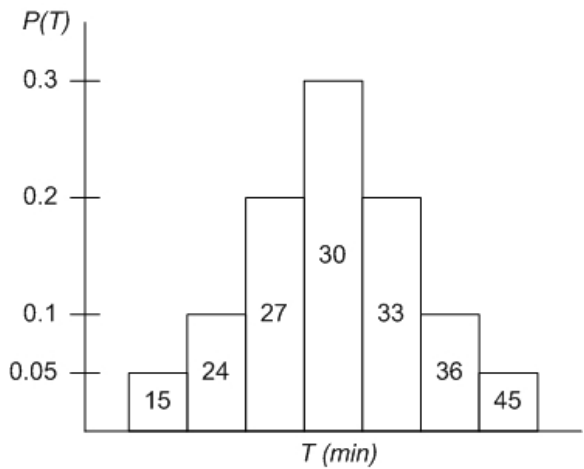

Figure 22: Distribution of load durations used for experiments.

For each load profile we ran best-of-two and the plan-based policy so that we could perform a direct comparison of lifetime achieved and number of switches performed. This resulted in 16 load-execution experiments. For the first two load profiles we restricted the best-of-two policy to switch at most every 5 minutes, so that the best-of-two policy and the plan-based policy switched a similar number of times in an entire run. Our simulation results suggest that the plan-based policy should switch no more than about 20 times, but our experiments reveal that the noise in the sensor data leads to errors in the estimation of the state of charge which cause the policy to switch more frequently than we would anticipate. Frequent switching indicates that the policy is responding to spurious artifacts in the sensed data and to the variability in the real behaviour of the batteries. We discuss this further in Section 8.

The plan-based policy was applied every 36 seconds (0.01 hours), reflecting the granularity of the plans and learned policy. We also ran an experiment in which the best-of-two policy was allowed to switch every 36 seconds, to ensure that the results we obtained were not biased by offering the plan-based policy a faster reaction time, to changes in the battery state of charge, than best-of-two.

We wanted to establish whether the plan-based policy can achieve similar lifetimes to the best-of-two policy with a lower numbers of switches. We also wished to confirm that it is better than the naive but simple policy of sequencing, in which the first battery is used until it is dead, and then the second battery is used. This should be obvious (the sequencing policy is much worse in simulation), but the observed behaviour of the plan-based policy is superficially similar to sequencing, since it favours mostly using one battery until it is heavily discharged before switching to the second battery for significant intervals, so we thought it useful to perform a physical comparison. In the case of a 2-battery setup sequencing involves only 1 switch (the minimum number of switches possible in the two battery case).

We ran 21 complete experiments in total. In all of the plots showing battery voltages during these experiments, the last lowest point on the battery voltage curves (the red and green curves) are the points at which the corresponding battery died.

Figure 23 shows the best-of-two policy running on the second load profile. The curves show the characteristic discharge/recovery pattern, separated by a step separation caused by the internal resistance of the battery (when the battery is recovering its voltage is open circuit, when it is loaded it is then reduced by the internal resistance).

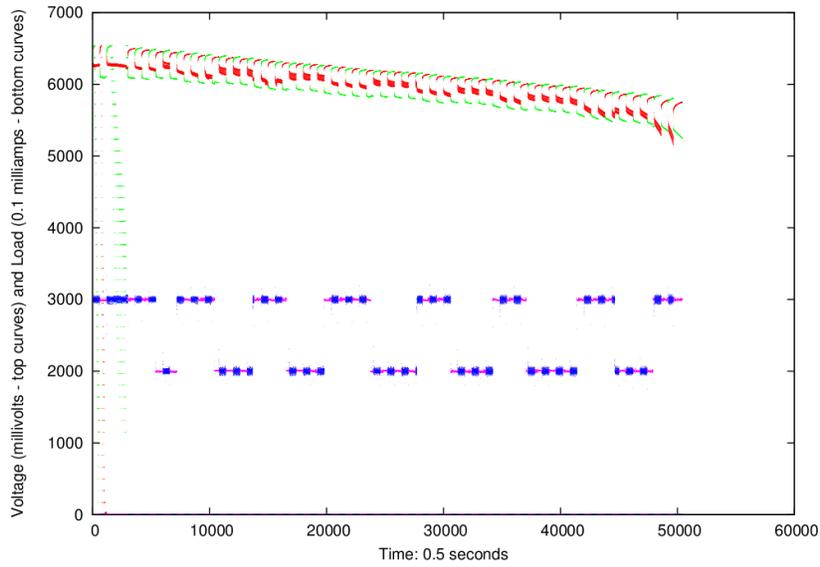

Figure 23: A run showing the behaviour of the best-of-two policy.

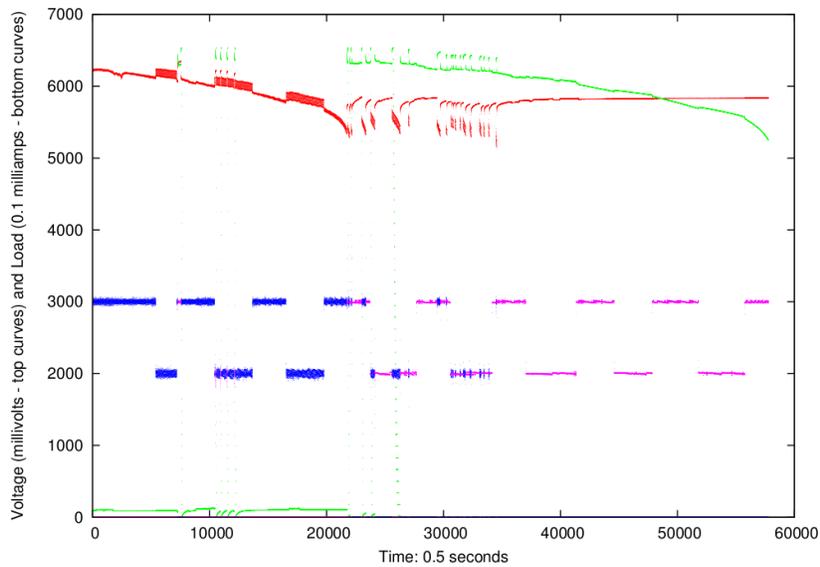

Figure 24: A run showing the behaviour of the plan-based policy.

The load and voltage curves for the red curve (battery $B_1$) are fuzzy because there is more noise in the readings from these sensors than for the other battery. This phenomenon is consistently a problem for $B_1$ and is not dependent on the battery, but appears to be a feature of the circuit itself.

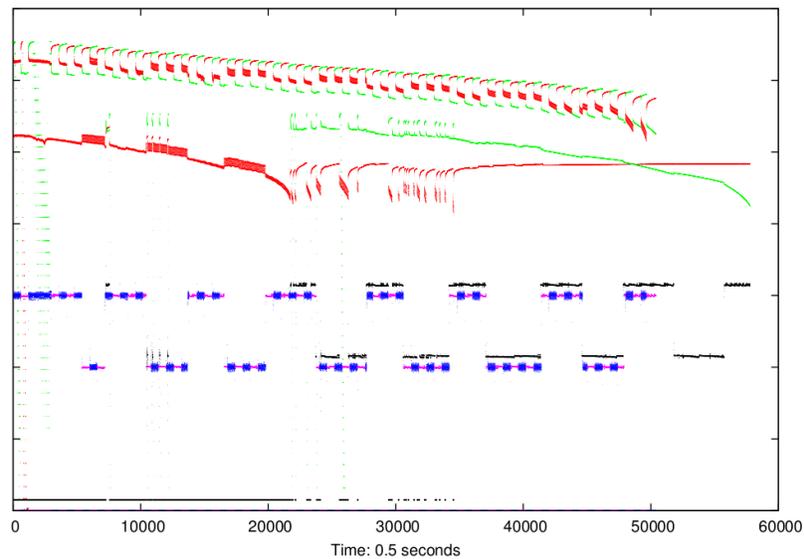

Time: 0.5 seconds

Figure 25: Best-of-two and the plan-based policy, both running on the same load profile. It can be seen that the lifetime achieved by the plan-based policy is longer, and the number of switches is also reduced. The y-axis has been removed, but load is measured in tenths of milliamps and voltage in tenths of millivolts, as before.

The strange striations for the green ($B_2$) curve at the start of the graph are due to a failure of the Arduino to correctly capture the battery voltage over this period, but it does not affect the performance of the policy (we have simple fail safes to ensure that spurious data of this sort do not affect our performance).

Figure 24 shows the behaviour of the plan-based policy running on the second load profile. The top two curves represent the usage of the two batteries, $B_1$ and $B_2$. Battery $B_1$ (the red curve) is used for the first 10,000 half-seconds, then $B_2$ is briefly used before the policy switches back to $B_1$ until about half way through the run. In the second half of the graph, the two batteries are interleaved, and the rising curves of $B_1$ correspond to the periods in which $B_2$ is in use and $B_1$ is resting.

The alternating load is represented by the bottom two curves. It can be seen that when the load changes, the measured voltage changes (the top curve registers a slight blip). This is because of the internal resistance which means that there is a lower voltage loss in the battery when the current changes. We would expect this to be about $34mV$ (if the internal resistance is $0.34\Omega$) because the difference in current is $0.1A$. It is actually higher than that, but this appears to be because there is a slight over-reaction to changes in the load, causing the battery voltage to drop sharply when the battery is first loaded, and then pull back, while the battery tends to recover sharply, and then fall back in line, when its load is reduced.

Figure 25 shows the best-of-two policy and the plan-based policy both being run on the second load profile side-by-side. The red plots are $B_1$ and green are $B_2$. The blue and purple points shows where $B_1/B_2$ serviced the load (and the value of the load) for best-of-two, while the black points, slightly displaced above these, show where $B_2$ serviced the load under the plan-based policy ($B_1$ serviced the load the rest of the time). The voltage curves for the plan-based policy have been offset from curves for best-of-two so that they can be displayed on the same plot. The labelling on the y-axis has been removed to avoid confusion. We can see three interesting features:

1. The plan-based policy tends to use $B_1$ first and $B_2$ second, although not sequentially.

2. The plan-based policy runs for longer, demonstrating that increased lifetime is achieved.

3. Best-of-two essentially alternates between the batteries (minor variations are due to slight discrepancies in the batteries and other factors).

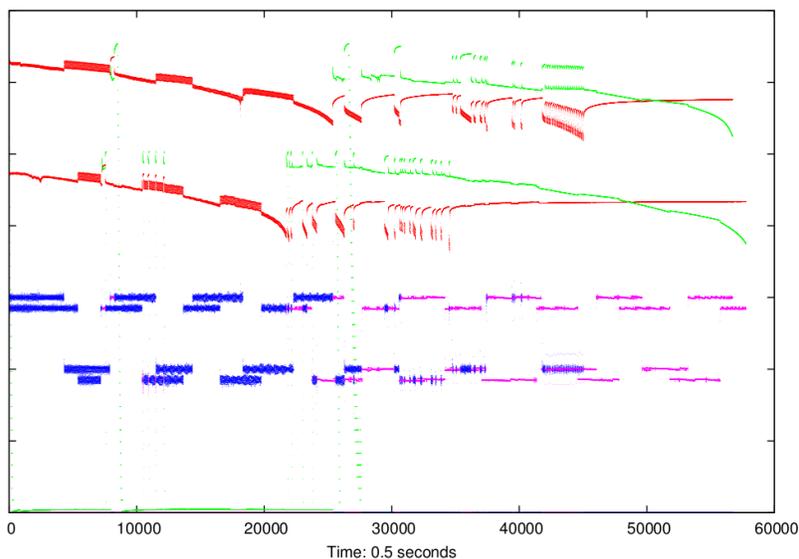

Figure 26: Two executions of the plan-based policy on different load profiles. The y-axis has been removed, but load is measured in tenths of milliamps and voltage in tenths of millivolts, as before.

Figure 26 shows a comparison of the plan-based policy working on the first and second load profiles. The performance of the policy on the first load profile is shown in the upper voltage curves and the upper load curves, while the curves for the second load profile have been displaced to differentiate them. The plot highlights the similarity in the way the policy manages the batteries in each case: the general strategy is to run $B_1$ until it is at the knee, resting it only briefly in this period, then oscillate between $B_1$ and $B_2$ at low frequency for a while, before entering a period in which $B_1$ is rapidly switched with $B_2$ as $B_1$ converges on empty. The policy then finishes off with $B_2$.

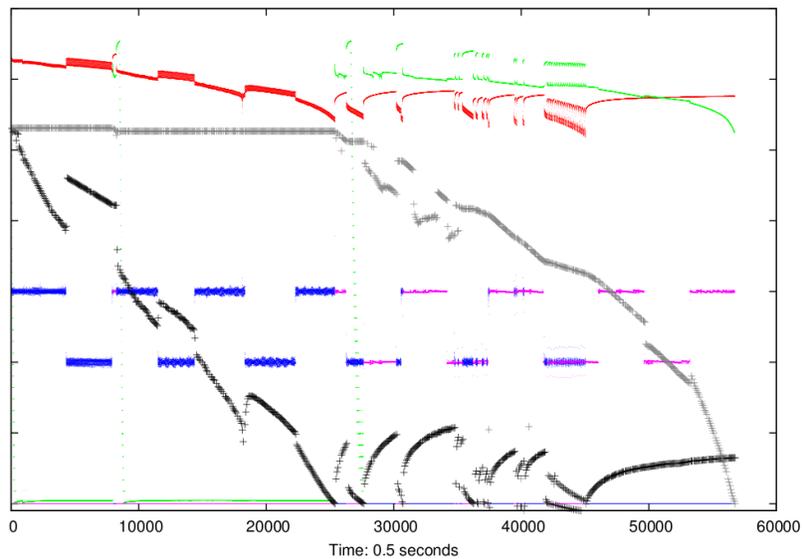

Figure 27: The plan-based policy plotted with estimated available charge. The y-axis has been removed, but load is measured in tenths of milliamps and voltage in tenths of millivolts, as before. Charge is measured in tenths of milliamp hours.

An interesting difference is a consequence of the (random) loads: $B_1$ is faced with heavier loads during the first part of the second profile, so it dies faster than in the first profile. However, $B_2$ faces a slightly less arduous time during the second half of the second profile and manages to last considerably longer. In particular, the load in the interval 30,000–33,000 was a high load serviced by $B_2$ in the first profile, while the same period happens to be a lower load in the second profile. This is a key reason why $B_2$ dies faster in the first profile: its available charge is depleted in that period and there is no real opportunity to rest it after that point. The final period of load in the first profile is a high load and that kills $B_2$ quickly, while the final period of load in the second profile is a lower one. This allows $B_2$ to recover some of its bound charge over that period, depleting its available charge more slowly and sustaining it a little longer in that critical period.

In Figure 26 the upper policy execution switches frequently in the window between 41,000 and 43,000 half seconds, just before $B_1$ dies. This is because the plan-based policy includes a default action to switch to the other battery to avoid the currently loaded battery dying prematurely. The reason for this is to protect the batteries and the policy from the effects of errors in the sensor data that propagate into the state of charge model. The effect of the default action in this case is to cause the policy to switch to $B_2$ when $B_1$ is almost out of charge, but back to $B_1$ as soon as it has recovered enough to be able to be loaded once again (according to the state of charge model).

Figure 27 shows the policy for the first load profile again, this time plotted with the estimated available charge (based on the voltage readings and the voltage model). The graph shows several important features. The black crosshairs mark the estimated available charge (measured in 0.1mAh units) for B1 and the grey crosshairs show it for B2. The discontinuities are due to the changing load values. There should be no discontinuity, because the model adjusts for the load (using our estimated internal resistance), but it is clear that there is an additional effect here that we cannot capture this way. As we have already mentioned, it is also the case that the discrepancy between battery terminal voltage readings for the different loads should be $0.1A \times 0.34\Omega = 34mV$, where $0.1A$ is the difference in load and $0.34\Omega$ is the internal resistance, but the graph shows differences that are much greater. This effect appears to worsen as the battery discharges (see the widening gaps between the loaded and unloaded voltages recorded for the batteries in the red/green curves — particularly for the red curve). However, interestingly, the voltage-capacity model seems to be marginally less unstable for lower states of charge (the steps get slightly smaller in these cases for the black curve).

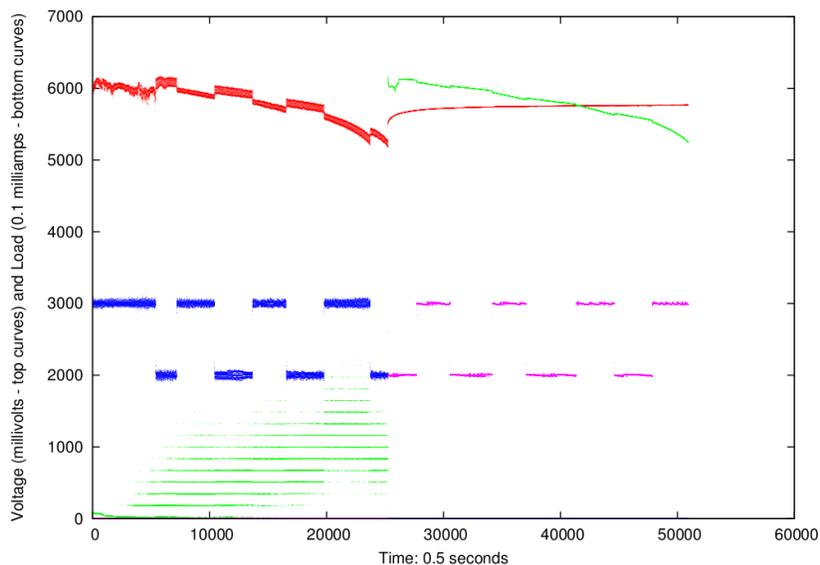

Figure 28: The sequencing policy showing its shorter lifetime on load profile 2.

As can also be seen, the available charge model breaks down in some situations (when the observations cannot be fitted consistently to the initial state we assumed for the battery). This leads to some of the available charge values being negative (particularly in the 42000–45000 period). This causes the policy to revert to the default action, but the somewhat simplistic implementation of the default leads to the oscillation between batteries during this period.

Figure 28 shows the results obtained by draining the batteries in sequence, using the second load profile. This performance is optimal in terms of switching, but the lifetime achieved is much shorter than that achieved by the plan-based policy and similar to the lifetime of the best-of-two for this case. The fact that best-of-two does worse than sequential scheduling for this profile is probably due to variation in the battery behaviour: it seems likely that best-of-two should perform more similarly to the results in the other load profiles.

It can be clearly seen that the plan-based policy achieves a consistently longer lifetime than the best-of-two policy, with significantly reduced switching. The results are summarised in Table 5.

| Load | Plan-based Policy | | Best-of-two | | Sequential | | Max. |
|------|----------|----------|----------|----------|----------|----------|------|
| Profile | Lifetime | Switches | Lifetime | Switches | Lifetime | Switches | |
| 1 | 7.887 | 71 | 7.534 | 73 | – | – | 8.77 |
| 2 | 8.033 | 47 | 7.000 | 81 | 7.079 | 1 | 8.91 |
| 3 | 7.974 | 91 | 7.563 | 705 | – | – | 9.04 |
| 4 | 7.831 | 158 | 6.998 | 701 | – | – | 9.23 |
| 5 | 7.030 | 17 | 6.226 | 609 | – | – | 9.11 |
| 6 | 7.120 | 36 | 7.085 | 706 | – | – | 8.81 |
| 7 | 7.669 | 21 | 7.645 | 649 | – | – | 9.11 |
| 8 | 7.677 | 88 | 6.515 | 584 | – | – | 8.87 |
| 9 | 8.341 | 33 | 5.901 | 567 | – | – | 8.91 |
| 10 | 6.972 | 13 | 6.890 | 690 | – | – | 8.92 |
| Mean | 7.653 | 57.5 | 6.936 | 651.4 | 7.079 | 1 | 8.97 |

Table 5: Table summarising results of physical experiments. Lifetimes are given in 'hours', but these are as reported using the Arduino clock and our measurements revealed that an hour measured by our Arduino was approximately 1.4 hours of real time. The first two experiments used lower switching frequency for the Best-of-two policy: as can be seen, the increased frequency for the later experiments does not offer any apparent advantage. These two results are not included in calculating the mean number of switches for the Best-of-two policy.

A paired $t$-test on these results shows that they are significant ($p = 0.013$). We expect that these improvements will be even more marked in the case of $n > 2$ batteries, but performing such experiments is the topic of future work. The final column in the table, labelled "max" shows the theoretical maximum lifetime of the batteries for the given load profile. These values are probably rather higher than the maximum value that could be achieved in practice, since the point at which the batteries are considered dead is based on observed terminal voltages when loaded. The internal resistance of the batteries means that this point is earlier than it is in the idealised battery model used in the simulation. The average efficiency of the batteries is 85% with our policy and 77% with the best-of-two compared with this theoretical maximum, which is consistent with both the expectation that the theoretical value is rather high and with previously reported performance of battery management systems that typically achieve around 80% efficiency.

## 8. Future Work

This paper brings together three distinct directions of research. Firstly, the work is concerned with a specific problem and its solution: the management of multiple batteries. Secondly, we develop and exploit techniques for planning with PDDL+ and continuous non-linear dynamics. Thirdly, we devise and implement an approach to policy construction based on planning for deterministic samples. Each of these directions offers scope for further work.

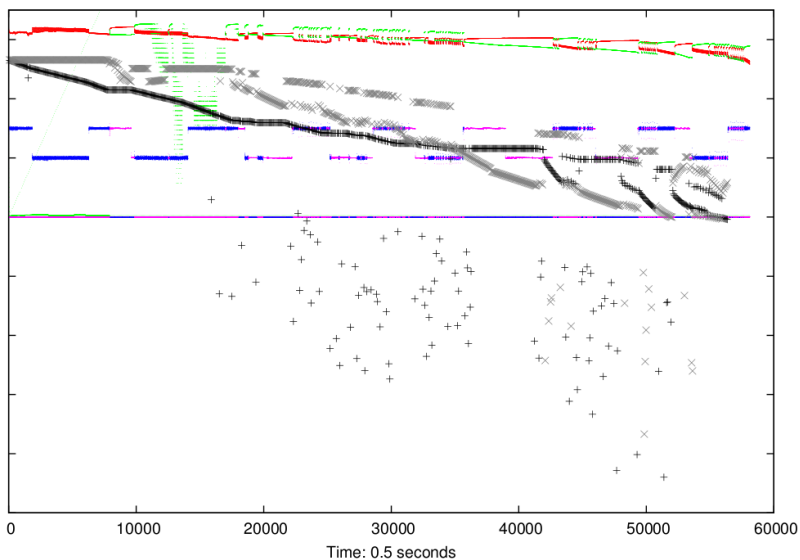

Figure 29: The battery voltage, load and estimated charge curves for the plan-based policy running on load profile 4. The y axis shows millivolts, 0.1 milliamps or 0.1 milliamp hours for each curve respectively.

The research on battery management has potential for real application and our physical experiments reveal that the theoretical results translate into measurable benefits. The physical experiments show higher switching rates for the plan-based policy control than our simulation results lead one to expect and we have noted that a key reason for this is the errors in the attempt to diagnose the state of charge of the batteries from noisy sensed voltage data. We anticipate that more robust sensing could resolve this problem to some extent, but a further modification is to consider a more careful implementation of the default action and of the tracking of state of charge. Figure 29 shows that in the plan-based policy run on the fourth load profile, the estimated available charge is often judged to be negative! This triggers application of the default action and in many cases these switches are contrary to the policy choices on either side of the spurious data point. In fact, of the 158 switches in this execution run, at least 90 are generated by spurious data triggering default actions. Similarly, for load profiles 1–3 we can identify at least 50, 8 and 54 cases respectively, in which the default action causes a switch in batteries against the advice of the policy for more sensible state of charge estimates on either side of the switches. This strongly suggests that a more careful implementation of the estimation of the state of charge, respecting the expected continuity of the behaviour, could lead to much better switching rates and better stability in the behaviour of the policy.

The experiments would obviously benefit from being performed on more a robustly constructed experimental apparatus and from additional runs to accumulate additional data. We hope to continue to pursue this direction in collaboration with commercial partners who might be interested in exploiting our ideas to achieve fielded systems.

The work on continuous planning, particularly for problems that include complex processes and events, remains a focus of research interest for us. We are now considering problems arising in different domains, including control of autonomous underwater vehicles and control of power systems (Bell, Coles, Coles, Fox, & Long, 2009). We are also exploring the ways in which hybrid planning might interface effectively with lower control levels through a shared model of system dynamics. The role of dynamic discretisation in managing complex process dynamics, particularly for non-linear behaviours, is one that we are continuing to explore.

Our work on the construction of policies via classification of trajectory samples built with a planner applied to sampled initial states is also a direction we are continuing to pursue. Our recent work on algal bloom mapping (Fox et al., 2012) indicates the directions we are considering. In particular, the states used in a planning model to allow a planner to solve sampled problem instances need not be the same as the states that are used in learning a policy. This is important, because the planner can exploit knowledge available in determinised instances of the problem to find high quality solutions and we can then hope that by careful selection of the observable elements of the visited states to be presented to the classifier, the classification process can discover correlations between the observable states and the actions selected by the planner in those states, in order to identify effective policy structures. This is a potentially powerful way to approach planning under uncertainty and we intend to investigate it much further.

## 9. Conclusions

This paper has presented an interesting and potentially important problem, managing systems powered by multiple independent batteries, and constructed a novel solution to it. In doing so we have brought together research on planning and policy learning to arrive at a new and powerful approach. We have experimentally evaluated our plans and learned policies in simulation and these results reveal that our solution can achieve better than 99% efficiency compared with the theoretical optimal (which is unachievable in practice). Not only do we achieve very high efficiencies, but we do so at low cost in terms of battery switching. This is beneficial because switching is wasteful of energy and tends to reduce the quality of service without additional smoothing circuitry that adds to energy losses.

Having confirmed our results in simulation we have gone on to explore the behaviour of the ideas in physical tests and those results confirm that real batteries are far less well-behaved than their simulated counterparts. Nevertheless, the policies we learn continue to behave very successfully — indeed we get results showing between 5% and 15% lifetime improvements over the best-of-two policy on equal load profiles, while still achieving lower switching rates.

Our approach to solving the battery usage problem adapts several existing technologies for automated planning, to solve a problem that can be seen as an MDP. We use Monte Carlo sampling to generate instances of determinised load profiles and solving these problems using an optimal deterministic solver, before combining the solutions to form a policy. Adopting a sampling approach to tackling problem-solving under uncertainty has become increasingly common and one of the reasons for this is that it usually offers better scaling opportunities than attempting to explicitly reason with distributions. Our policy construction approach adapts the use of machine learning to construct a classifier. In the construction of high quality solutions to deterministic problems, we use a special variable-range discretisation to solve a non-linear continuous optimisation problem with very high accuracy, while exploring a very small proportion of the state space.

Our approach is scalable and effective. Although the solution as we implement it for this paper is domain-specific in several respects, the components are general and we have already begun to illustrate this point by adapting the approach to other problems. The elements that are most tailored to our problem are the selection of the discretisation range and the search heuristic. However, we believe that the characteristics of the multiple battery usage problem are shared, in outline, by other domains and expect the approach can be adapted to these domains with relative ease.

## Acknowledgments

We would like to thank Marijn Jongerden and Boudewijn Haverkort for introducing us to the multiple battery usage problem, and drawing our attention to the scheduling problem and related policy-based approaches. We would also like to extend our thanks to the anonymous reviewers and the handling editor, Carmel Domshlak, for their help in improving the text of the paper.

This work was partially funded by the EPSRC Project "Automated Modelling and Reformulation in Planning" (EP/G0233650).